\documentclass[runningheads]{llncs}

\usepackage[mobile]{eccv}

\usepackage{eccvabbrv}

\usepackage{graphicx}
\usepackage{booktabs}

\usepackage{caption}
\usepackage{subcaption}
\usepackage{amsmath}
\usepackage{amssymb}
\usepackage{enumitem}

\usepackage{bm}
\usepackage{listings}
\usepackage{makecell}  % \Xhline
\usepackage{diagbox} % diagonal table cell
\usepackage{mathtools} % \coloneqq :=
\usepackage{graphicx} % rotated text
\usepackage{multirow}

\usepackage{mathrsfs}

\usepackage{xcolor}
\usepackage{caption}
\usepackage{subcaption}
\usepackage{tabularx}
\usepackage{rotating}
\usepackage{verbatim}
\usepackage{xcolor,colortbl}
\usepackage{etoolbox}
\usepackage[normalem]{ulem}
\usepackage{booktabs}
\usepackage{multirow}
\usepackage{lipsum}
\usepackage{stmaryrd}
\usepackage{stackengine}
\usepackage{makecell}
\usepackage{pifont}
\usepackage{cancel}
\usepackage{adjustbox}
\usepackage{dblfloatfix}
\usepackage{bbding}
\usepackage{pifont}
\usepackage{wasysym}
\usepackage{amssymb}

\usepackage{algorithm}
\usepackage{algpseudocode}
\usepackage{amsmath}
\makeatletter
\newcommand{\car@semkitfreq}{3.92}
\newcommand{\bicycle@semkitfreq}{0.03}
\newcommand{\motorcycle@semkitfreq}{0.03}
\newcommand{\truck@semkitfreq}{0.16}
\newcommand{\othervehicle@semkitfreq}{0.20}
\newcommand{\person@semkitfreq}{0.07}
\newcommand{\bicyclist@semkitfreq}{0.07}
\newcommand{\motorcyclist@semkitfreq}{0.05}
\newcommand{\road@semkitfreq}{15.30}  %
\newcommand{\parking@semkitfreq}{1.12}
\newcommand{\sidewalk@semkitfreq}{11.13}  %
\newcommand{\otherground@semkitfreq}{0.56}
\newcommand{\building@semkitfreq}{14.1}  %
\newcommand{\fence@semkitfreq}{3.90}
\newcommand{\vegetation@semkitfreq}{39.3}  %
\newcommand{\trunk@semkitfreq}{0.51}
\newcommand{\terrain@semkitfreq}{9.17} %
\newcommand{\pole@semkitfreq}{0.29}
\newcommand{\trafficsign@semkitfreq}{0.08}
\newcommand{\semkitfreq}[1]{{\csname #1@semkitfreq\endcsname}}

\definecolor{White}{rgb}{1.,0.,1.}
\definecolor{first}{rgb}{.8,.0,.0}
\definecolor{second}{rgb}{.0,.6,.0}
\definecolor{third}{rgb}{.0,.0,.8}
\newcolumntype{g}{>{\columncolor{White}}c}

\definecolor{car.}{rgb}{0.39215686, 0.58823529, 0.96078431}
\definecolor{bicycle.}{rgb}{0.39215686, 0.90196078, 0.96078431}
\definecolor{motorcycle.}{rgb}{0.11764706, 0.23529412, 0.58823529}
\definecolor{truck.}{rgb}{0.31372549, 0.11764706, 0.70588235}
\definecolor{othervehicle.}{rgb}{0.39215686, 0.31372549, 0.98039216}
\definecolor{person.}{rgb}{1.        , 0.11764706, 0.11764706}
\definecolor{bicyclist.}{rgb}{1.        , 0.15686275, 0.78431373}
\definecolor{motorcyclist.}{rgb}{0.58823529, 0.11764706, 0.35294118}
\definecolor{road.}{rgb}{1.        , 0.        , 1.        }
\definecolor{parking.}{rgb}{1.        , 0.58823529, 1.        }
\definecolor{sidewalk.}{rgb}{0.29411765, 0.        , 0.29411765}
\definecolor{otherground.}{rgb}{0.68627451, 0.        , 0.29411765}
\definecolor{building.}{rgb}{1.        , 0.78431373, 0.        }
\definecolor{fence.}{rgb}{1.        , 0.47058824, 0.19607843}
\definecolor{vegetation.}{rgb}{0.        , 0.68627451, 0.        }
\definecolor{trunk.}{rgb}{0.52941176, 0.23529412, 0.        }
\definecolor{terrain.}{rgb}{0.58823529, 0.94117647, 0.31372549}
\definecolor{pole.}{rgb}{1.        , 0.94117647, 0.58823529}
\definecolor{trafficsign.}{rgb}{1.        , 0.        , 0.    }

\definecolor{detcolor}{gray}{.9}

\definecolor{bestcolor}{gray}{.9}
\newcommand{\bestcell}[1]{\cellcolor{bestcolor}{#1}}

\newlength\savewidth
  
\newcolumntype{x}[1]{>{\centering\arraybackslash}p{#1pt}}
\newcolumntype{y}[1]{>{\raggedright\arraybackslash}p{#1pt}}
\newcolumntype{z}[1]{>{\raggedleft\arraybackslash}p{#1pt}}
\renewcommand{\paragraph}[1]{\vspace{1.25mm}\noindent\textbf{#1}}
\definecolor{deemph}{gray}{0.6}

\usepackage{etoolbox}
\makeatletter
\AfterEndEnvironment{algorithm}{\let\@algcomment\relax}
\AtEndEnvironment{algorithm}{\kern2pt\hrule\relax\vskip3pt\@algcomment}
\let\@algcomment\relax
\newcommand\algcomment[1]{\def\@algcomment{\footnotesize#1}}
\renewcommand\fs@ruled{\def\@fs@cfont{\bfseries}\let\@fs@capt\floatc@ruled
  \def\@fs@pre{\hrule height.8pt depth0pt \kern2pt}%
  \def\@fs@post{}%
  \def\@fs@mid{\kern2pt\hrule\kern2pt}%
  \let\@fs@iftopcapt\iftrue}
\makeatother

\definecolor{barrier}{RGB}{112,128,144}
\definecolor{bicycle}{RGB}{220,20,60}
\definecolor{bus}{RGB}{255, 127, 80}
\definecolor{car}{RGB}{255, 158, 0}
\definecolor{const. veh.}{RGB}{233, 150, 70}
\definecolor{motorcycle}{RGB}{255,61,99}
\definecolor{pedestrian}{RGB}{0,0,230}
\definecolor{traffic cone}{RGB}{47,79,79}
\definecolor{trailer}{RGB}{255,140,0}
\definecolor{truck}{RGB}{255,99,71}
\definecolor{drive. suf.}{RGB}{0,207,191}
\definecolor{other flat}{RGB}{175,0,75}
\definecolor{sidewalk}{RGB}{75,0,75}
\definecolor{terrain}{RGB}{112,180,60}
\definecolor{manmade}{RGB}{222,184,135}
\definecolor{vegetation}{RGB}{0,175,0}

\definecolor{turquoise}{cmyk}{0.65,0,0.1,0.1}
\definecolor{purple}{rgb}{0.65,0,0.65}
\definecolor{dark_green}{rgb}{0, 0.5, 0}
\definecolor{orange}{rgb}{0.98, 0.51, 0}
\definecolor{red}{rgb}{1.0, 0.1, 0.1}
\definecolor{blue}{rgb}{0.2, 0.2, 1.0}
\definecolor{brown}{rgb}{0.5, 0.16, 0.16}

\newcommand{\red}[1]{{\color{red} \textbf{#1}}}
\newcommand{\blue}[1]{{\color{blue} \textbf{#1}}}

\newcommand{\cmark}{\ding{51}}%
\newcommand{\xmark}{\ding{55}}%

\usepackage[accsupp]{axessibility}  % Improves PDF readability for those with disabilities.

\usepackage[pagebackref,breaklinks,colorlinks,citecolor=eccvblue]{hyperref}
\usepackage{orcidlink}

\begin{document}

% ---------------------------------------------------------------
% TODO REVIEW: Replace with your title
\title{OccGen: Generative Multi-modal 3D Occupancy Prediction for Autonomous Driving} 

% TODO REVIEW: If the paper title is too long for the running head, you can set
% an abbreviated paper title here. If not, comment out.
\titlerunning{OccGen}

% TODO FINAL: Replace with your author list. 
% Include the authors' OCRID for the camera-ready version, if at all possible.
% \author{Guoqing Wang\inst{1} \and
% Zhongdao Wang\inst{2,3}\orcidlink{1111-2222-3333-4444} \and
% Third Author\inst{3}\orcidlink{2222--3333-4444-5555}}

\author{Guoqing Wang\inst{1} 
\and Zhongdao Wang\inst{2} 
\and Pin Tang\inst{1} 
\and Jilai Zheng\inst{1} 
\and Xiangxuan Ren\inst{1} 
\and Bailan Feng\inst{2} 
\and Chao Ma\inst{1}\thanks{Corresponding author}
}
\authorrunning{G.~Wang, Z.~Wang et al.}
\institute{$^1$MoE Key Lab of Artificial Intelligence, AI Institute, Shanghai Jiao Tong University, $^2$Huawei Noah's Ark Lab
{\tt\small \{guoqing.wang,pin.tang,zhengjilai,bunny\_renxiangxuan,chaoma\}@sjtu.edu.cn} \\
{\tt\small \{wangzhongdao,fengbailan\}@huawei.com}\\
{\small Project page: \url{https://occgen-ad.github.io/}}
}

% TODO FINAL: Replace with an abbreviated list of authors.
% \authorrunning{F.~Author et al.}
% First names are abbreviated in the running head.
% If there are more than two authors, 'et al.' is used.

% TODO FINAL: Replace with your institution list.

\maketitle

\begin{abstract}
  Existing solutions for 3D semantic occupancy prediction typically treat the task as a one-shot 3D voxel-wise segmentation perception problem. These discriminative methods focus on learning the mapping between the inputs and occupancy map in a single step, lacking the ability to gradually refine the occupancy map and the reasonable scene imaginative capacity to complete the local regions somewhere. 
  In this paper, we introduce OccGen, a simple yet powerful generative perception model for the task of 3D semantic occupancy prediction. OccGen adopts a ``noise-to-occupancy'' generative paradigm, progressively inferring and refining the occupancy map by predicting and eliminating noise originating from a random Gaussian distribution. OccGen consists of two main components: a conditional encoder that is capable of processing multi-modal inputs, and a progressive refinement decoder that applies diffusion denoising using the multi-modal features as conditions.  A key insight of this generative pipeline is that the diffusion denoising process is naturally able to model the coarse-to-fine refinement of the dense 3D occupancy map, therefore producing more detailed predictions. Extensive experiments on several occupancy benchmarks demonstrate the effectiveness of the proposed method compared to the state-of-the-art methods. For instance,  OccGen relatively enhances the mIoU by  9.5\%, 6.3\%, and 13.3\% on nuScenes-Occupancy dataset under the muli-modal, LiDAR-only, and camera-only settings, respectively. Moreover, as a generative perception model, OccGen exhibits desirable properties that discriminative models cannot achieve, such as providing uncertainty estimates alongside its multiple-step predictions.
\end{abstract}

\section{Introduction}
\label{sec:intro}

% \vspace{-1.0em}
\begin{figure}[t]
\begin{center}
%\fbox{\rule{0pt}{2in} \rule{0.9\linewidth}{0pt}}
\includegraphics[width=0.95\linewidth]{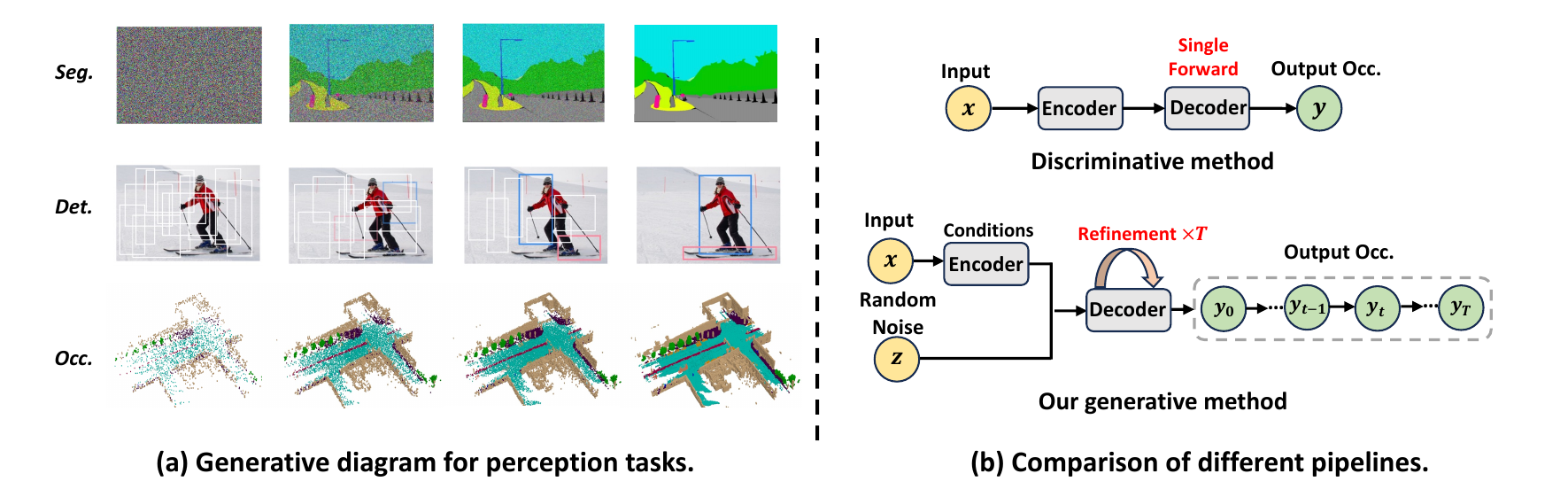}
\vspace{-2.0em}
\end{center}
   \caption{
   (a) The generative diagram of semantic segmentation (seg.), object detection (det.), and 3D semantic occupancy prediction (occ.).
   (b) Compared to previous discriminative methods with a single forward evaluation scheme, our OccGen is a generative model that can generate occupancy map in a coarse-to-fine manner.}
   
   % (b) A comparison against the state-of-the-art CONet \cite{wang2023openoccupancy} at different sampling steps of OccGen on nuScenes-Occupancy. OccGen achieves higher performance with an increasing number of sampling steps. The relative gains are marked in red.
   % \zd{Figure 1(b) seems a bit weak, put the performance here may not be wise. Instead it may be better to show some visualizations of the diffusion process, as done in the DDP paper, Figure 1.}}
\label{fig_intro}
\vspace{-1.5em}
\end{figure}

% 从3D object detection引出3D Occupancy prediction
The precise 3D perception of the surrounding environment constitutes the cornerstone of modern autonomous driving systems, as it directly affects downstream tasks such as planning and vehicle control \cite{hu2023planning, jia2023driveadapter}. 
In recent years, advancements in 3D object detection and segmentation
\cite{Lang2019PointPillarsFE, Zhou2018VoxelNetEL, bevdet, li2022bevdepth, li2022bevformer, zhu2021cylindrical, zhang2020polarnet, liu2022bevfusion, liang2022bevfusion, Vora2020PointPaintingSF, Wang2021PointAugmentingCA, lu2024scaling, lu2023towards} have propelled the field of 3D perception.
% These approaches primarily depend on two types of perception sensors \cite{caesar2020nuscenes, geiger2012we, Sun2020ScalabilityIP}, i.e., LiDAR and camera.  
% LiDAR points can offer precise accurate spatial location and depth information for specific scenes, yet their performance is constrained by inherent limitations of sparsity and lack of texture. Conversely, 2D images exhibit dense semantic information, vibrant color, and subtle textures, but they lack the inherent depth information found in LiDAR points. Recently, LiDAR-camera fusion methods \cite{liu2022bevfusion, liang2022bevfusion, Vora2020PointPaintingSF, Wang2021PointAugmentingCA} have been designed to leverage information from both modalities and have shown promising results in various benchmarks.
However, these methods require either rigid bounding boxes, which oversimplify the object shapes, or Bird's-Eye View (BEV) predictions that involve compromises in projecting 3D scenes onto 2D ground planes. Such methods can significantly impede the ability to accurately perceive structural information along the vertical axis, particularly when dealing with irregular objects. 

To address this limitation, 3D semantic occupancy prediction~\cite{song2017semantic, huang2023tri, wei2023surroundocc, tong2023scene, tian2023occ3d} has been proposed to assign semantic labels to every spatially occupied region within the perceptive range. 
Most previous methods for 3D semantic occupancy prediction can be roughly divided into three categories: LiDAR-based \cite{lmscnet, js3cnet, sketch, aicnet}, vision-based \cite{cao2022monoscene, li2023voxformer, huang2023tri, Zhang_2023_ICCV, tang2024sparseocc}, and multi-modal based~\cite{wang2023openoccupancy} methods. These methods typically formulate the 3D occupancy prediction as a one-shot voxel-wise segmentation problem with a single forward evaluation scheme.
% The LiDAR-based methods \cite{lmscnet, js3cnet, sketch, aicnet} are proposed to fill in gaps or areas where LiDAR sweeps might be sparse or missing in a 3D environmental scene. Vision-based perception methods \cite{cao2022monoscene, li2023voxformer, huang2023tri, Zhang_2023_ICCV} recently emerged as a promising alternative to LiDAR-based one to effectively lift 2D camera features to 3D voxel space for the subsequent semantic scene completion. However, such a projection inevitably assigns 2D features of visible regions to the empty or occluded voxels.
% The pioneering work OpenOccupancy \cite{wang2023openoccupancy} constructed the first 3D multi-modal occupancy prediction benchmark and proposed an effective CONet to alleviate the computational burden of high-resolution occupancy predictions.
While these works achieve promising results, this perception pipeline faces two critical issues: 1) Discriminative methods primarily focus on learning the mapping between the input-output pairs in a single forward step and neglect the modeling of the underlying occupancy map distribution. 2) Inferring only once is not enough for the model to complete the fine-grained scene well, just like humans need continuous observation to fully perceive the entire scene.
% they still focus on learning the manifold between input-output pairs in a single forward step without modeling the underlying distribution of the occupancy.
% This limitation leads to poor generalization to envision the entire 3D geometry of occluded objects and scenes.
% \zd{Same problem, too much introduction of the background. Need to cut to the chase as soon as possible.}

On the other hand, the diffusion model~\cite{ho2020denoising, song2020denoising} has demonstrated its powerful generation capability and has also led to the successful application in numerous discriminative tasks, such as depth estimation \cite{saxena2023monocular, ji2023ddp}, object detection \cite{chen2023diffusiondet}, and segmentation \cite{wu2022medsegdiff,amit2021segdiff,wolleb2022diffusion}. We observe that the diffusion denoising process is naturally able to model the coarse-to-fine refinement of the dense 3D occupancy map, therefore producing more detailed predictions.
Motivated by this, we propose OccGen, a simple yet powerful generative perception model for 3D semantic occupancy. As shown in Fig.~\ref{fig_intro}, OccGen adopts a ``noise-to-occupancy'' generative paradigm, progressively inferring and eliminating noise originating from a random 3D Gaussian distribution. The proposed OccGen consists of two main components: a conditional encoder and a progressive refinement decoder. The conditional encoder only needs to run once, while the decoder runs multiple times to fulfill progressive refinement. Since the encoder only runs once during the entire inference process, running the decoder step-by-step for diffusion denoising does not introduce significant computational overhead, thereby achieving comparable latency to single forward methods.
During the training phase, we obtain a 3D noise map by gradually adding Gaussian noise to the ground truth occupancy. Subsequently, this noise map is fed into the progressive refinement decoder, which utilizes the multi-scale fusion features from the conditional encoder as conditions to generate noise-free predictions. In the inference phase, OccGen progressively generates the occupancy in a coarse-to-fine refinement manner, which is implemented by gradually denoising a 3D Gaussian noise map given the multi-modal condition inputs.

% At the training stage, we obtain noisy occupancy by gradually adding Gaussian noise to ground truth occupancy. Then, the noisy occupancy is sent to the progressive refinement module to produce the predictions without noise regarding the multi-modal fusion features from the condition encoder as condition. At the inference stage, OccGen progressively generates occupancy by reversing the diffusion process, which adjusts a noisy Gaussian distribution to the learned occupancy distribution under the guidance of the multi-modal inputs.
As a generative perception model, OccGen exhibits desirable properties that are not achievable by discriminative models: (1) progressive inference supports trading compute for prediction quality; (2) uncertainty estimation can be readily made alongside the predictions. 
We evaluate the effectiveness of OccGen on several benchmarks and show promising results compared with the state-of-the-art methods. Notably, OccGen has exhibited performance gains of 9.5\%, 6.3\%, and 13.3\% on mIoU compared with the state-of-the-art method under the muli-modal, LiDAR-only, and camera-only settings on nuScenes-Occupancy. 

Our contributions are summarized as follows: 
\begin{itemize}
  \item We introduce OccGen, a simple yet powerful generative framework following the ``noise-to-occupancy'' paradigm.
  \item OccGen adapts an efficient design that the encoder only runs once during the entire inference process, and the decoder runs step-by-step for progressive refinement, achieving a comparable latency to single forward methods.
  \item We extensively validate the proposed OccGen on multiple occupancy benchmarks, demonstrating its remarkable performance and desirable properties compared to previous discriminative methods.
\end{itemize}

% DDP decouples the image encoder and
% map decoder. The image encoder runs only once, while the
% diffusion process is performed only in the lightweight decoder head. With this efficient design, our proposed method
% can easily be applied to modern perception tasks.

\section{Related Work}
\label{sec:related}

\vspace{-0.5em}
\subsection{3D Semantic Occupancy Prediction.}

The majority of popular 3D perception methods, whether the input is LiDAR sweeps, multi-view images, or multi-modal data, construct BEV feature representations and subsequently perform various downstream tasks in the BEV space \cite{bevdet, li2022bevdepth, li2022bevformer, liu2022bevfusion, liang2022bevfusion}. However, these BEV-based methods typically project the 3D scene onto the ground plane, leading to the potential loss of information in the vertical dimension. Compared with BEV representation, 3D semantic occupancy provides a more detailed representation of the environment by explicitly modeling the occupancy status of each voxel in a 3D grid.

SSCNet \cite{song2017semantic} has first introduced the task of semantic scene completion, integrating both geometry and semantics. Subsequent works commonly utilized geometric inputs with explicit depth information \cite{lmscnet, aicnet, js3cnet, sketch}. MonoScene \cite{cao2022monoscene} has proposed the first monocular approach for semantic scene completion, employing a 3D UNet \cite{ronneberger2015u} to process voxel features generated through sight projection. TPVFormer \cite{huang2023tri} proposed a tri-perspective view representation for describing 3D scenes in semantic occupancy prediction. VoxFormer \cite{li2023voxformer} introduced a two-stage transformer-based semantic scene completion framework that can output complete 3D volumetric semantics from only 2D images. OccFormer \cite{Zhang_2023_ICCV} introduced a dual-path transformer network for effective processing of 3D volumes in semantic occupancy prediction, achieving long-range, dynamic, and efficient encoding of camera-generated 3D voxel features. Furthermore, many concurrent works are dedicated to proposing surrounding-view benchmarks for 3D semantic occupancy prediction, contributing to the flourishing of the occupancy community \cite{wang2023openoccupancy, wei2023surroundocc, tong2023scene, tian2023occ3d}. OpenOccupancy \cite{wang2023openoccupancy} constructed the first 3D multi-modal occupancy prediction benchmark and proposed an effective CONet to alleviate the computational burden of high-resolution occupancy predictions. In this paper, we propose OccGen, a simple yet powerful generative perception framework for 3D multi-modal semantic occupancy that can progressively refine the occupancy in a coarse-to-fine manner. 

\vspace{-0.5em}
\subsection{Diffusion Model}
%TODO
% Diffusion [35, 71] and score-based generative models [73] have been particularly successful as generative models and achieve impressive results across
% various modalities, including images [60, 66, 27, 57, 25, 25], video [36, 37], audio [43], and biomedical [2, 77, 69, 22]. Given the notable achievements of diffusion models in these respective domains, leveraging such models to develop generation-based perceptual models would prove to
% be a highly promising avenue to push the boundaries of perceptual tasks to newer heights.

% With the recent success of diffusion models in generation tasks, there has
% been a noticeable rise in interest to incorporate them into dense visual prediction tasks. Several pioneering works [82, 1, 81, 14, 68, 12] attempted to apply the diffusion model to visual perception tasks, e.g. image segmentation or depth estimation task. For example, Wolleb et al. [81] explore the diffusion model for medical image segmentation. Pix2SeqD [14] applies the bit diffusion model [16] for panoptic segmentation. Our concurrent work DepthGen [68] involves
% diffusion pipeline to the task of depth estimation. For all the diffusion models listed above, one or two parameterheavy convolutional U-Nets [64] are adopted, leading to low efficiency, slow convergence, and sub-optimal performance.

Diffusion models \cite{sohl2015deep, ho2020denoising} have been extensively researched due to their powerful generation capability. Denoising diffusion probabilistic models (DDPM) \cite{ho2020denoising} proposed a diffusion model where the forward and reverse processes exhibit the Markovian property. Denoising diffusion
implicit models (DDIM) \cite{song2020denoising} accelerated DDPM \cite{ho2020denoising} by replacing the original diffusion process with non-Markovian chains to enhance sampling speed. On the other hand, conditional diffusion models have also been actively studied. Text-to-image generation models \cite{rombach2022high} and image-to-image translation models \cite{saharia2022palette} achieved surprising results. 
Recently, diffusion models for visual perception have attracted widespread attention. Several pioneering works \cite{wu2022medsegdiff,amit2021segdiff,wolleb2022diffusion, chen2022generalist,saxena2023depthgen} attempted to apply the diffusion model to visual perception tasks, e.g. image segmentation or depth
estimation tasks. For all the diffusion models listed above, one or two parameter-heavy convolutional U-Nets \cite{ronneberger2015u} are adopted, leading to low efficiency, slow convergence, and sub-optimal performance. DiffusionDet \cite{chen2023diffusiondet} 
proposed a denoising diffusion process from noisy boxes to object boxes.
DDP \cite{ji2023ddp} followed the ``noise-to-map'' generative paradigm for prediction by progressively removing noise from a random Gaussian distribution, guided by the image. In this work, as illustrated in Fig.~\ref{Fig_method}, we extend the generative diffusion process into the occupancy perception pipeline while maintaining accuracy and efficiency.

\section{Method}
\label{sec:method}
In this section, we first introduce the preliminaries on 3D semantic occupancy perception and conditional diffusion model. Then, we present the pipeline of the ``noise-to-occupancy'' and the overall architecture of OccGen. Finally, we show the details of the training and inference process.

\subsection{Preliminaries}

%重新画图
\begin{figure*}[t]
\begin{center}
%\fbox{\rule{0pt}{2in} \rule{0.9\linewidth}{0pt}}
\includegraphics[width=0.9\textwidth]{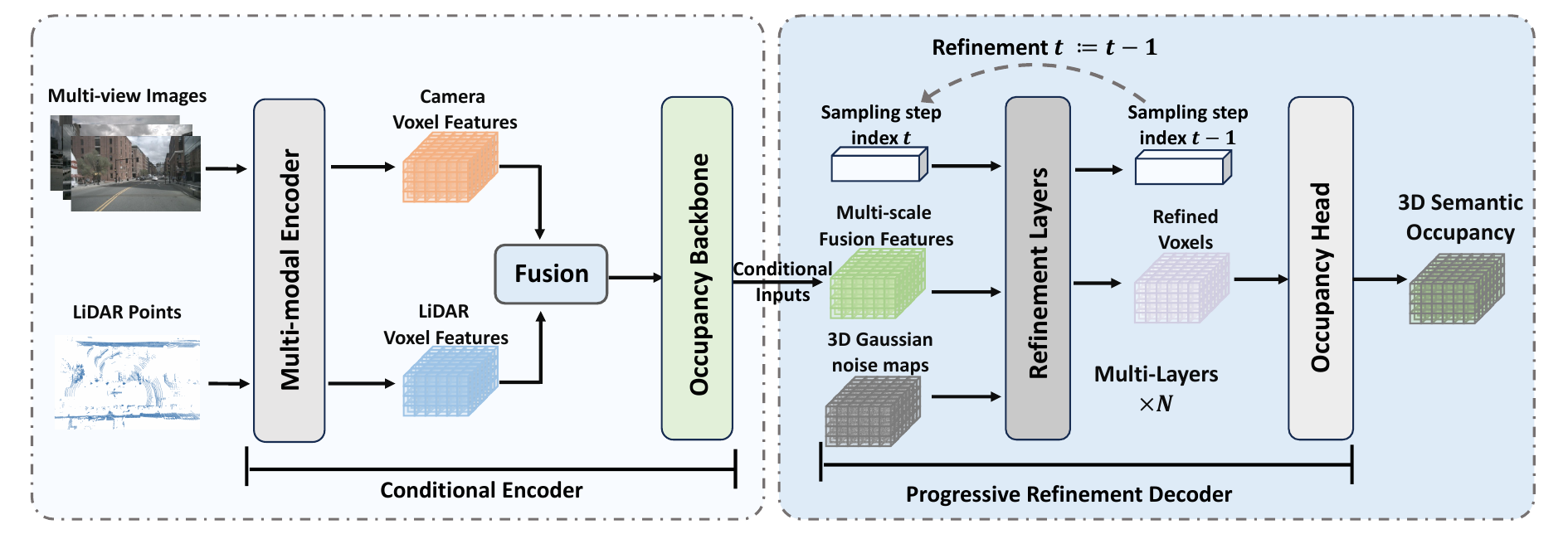}
\vspace{-1.3em}
\end{center}
   \caption{The proposed OccGen framework. It has an encoder-decoder structure. The conditional encoder extracts the features from the inputs as the condition. The progressive refinement decoder consists of a stack of refinement layers and an occupancy head, which takes the 3D noise map, sampling step, and conditional multi-scale fusion features as inputs and progressively generates the occupancy prediction.}
\label{Fig_method}
\vspace{-1.0em}
\end{figure*}

\paragraph{3D semantic occupancy perception.}
The objective of 3D semantic occupancy perception is to predict a complete 3D representation of volumetric occupancy and semantic labels for a scene in the surround-view driving scenarios given inputs, such as images and LiDAR points. We utilize LiDAR point cloud $X_{p} \in \mathbb{R}^{N_{L}\times(3+d)}$ and multi-view camera images $X_{c} \in \mathbb{R}^{N_{C}\times H_{C} \times W_{C} \times 3}$ as multi-modal inputs, denoted by $X = \{X_{p}, X_{c}\}$. Subsequently, we train a neural network $f_{\theta}$ to generate an occupancy voxel map $Y \in \{c_{0}, c_{1}, ..., c_{N}\}^{H \times W \times Z}$, where each voxel is assigned either an empty label $c_{0}$ or occupied by a specific semantic class from $\{c_{1}, c_{2}, ..., c_{N}\}$. Here, 
$N$ represents the total number of interested classes, and $\{H, W, D\}$ indicates the volumetric dimensions of the entire scene.
% \vspace{-0.5em}

\paragraph{Diffusion model.}
The diffusion model is a type of generative model that demonstrates greater potential in the generative domain compared to Generative Adversarial Network (GAN) \cite{goodfellow2014generative}. It can be divided into two categories: unconditional diffusion models learn an explicit approximation of the data distribution $P(z)$, while conditional diffusion models learn the distribution given a certain condition $k$, denoted as $p(z|k)$.
In the conditional diffusion model, the data distribution is learned by recovering a data sample from Gaussian noise through an iterative denoising process. The forward diffusion process gradually adds noise to the data sample $z_{0}$, denoted as:
\begin{equation}
\label{eq:forward_ddpm}
    z_t = \sqrt{{\alpha_t}}z_0  + \sqrt{1 - {\alpha_t}}{Z},\quad Z \sim \mathcal{N} (0,I),
\end{equation}
which transforms the $z_{0}$ to a latent noisy sample $z_{t}$ for $t \in\{0,1, \ldots, T\}$. The constant $\alpha_t = \prod_{i=1}^{t}(1 - \beta_{i})$ and $\beta_s$ represents the noise schedule. In the training process, the conditional diffusion model $f_\theta \left(\boldsymbol{z}_t, t\right|k)$ is trained to predict $\boldsymbol{z}_0$ from $\boldsymbol{z}_t$ under the guidance of condition $\boldsymbol{k}$ by minimizing the training objective function (\ie, $l_2$ loss). In the inference process, the predicted sample $\boldsymbol{z}_0$ is reconstructed from a random noise $\boldsymbol{z}_T$ with the model $f_\theta$ and conditional input $\boldsymbol{k}$ following the denoising process of DDPM \cite{ho2020denoising} or DDIM \cite{song2020denoising}.

% In this paper, our goal is to model the coarse-to-fine refinement of the dense 3D occupancy map via the conditional diffusion model. In our setting, the data samples and conditions are the ground truth occupancy $z_0 = Y$, and multi-modal features $k = F_M$, respectively. Then, the proposed progressive refinement module is trained to predict $z_0$ from random noise $z_t \sim \mathcal{N} (0,I)$ conditioned on the corresponding multi-modal features $F_M$.

\begin{figure*}[t]
\begin{center}
%\fbox{\rule{0pt}{2in} \rule{0.9\linewidth}{0pt}}
\includegraphics[width=1.0\textwidth]{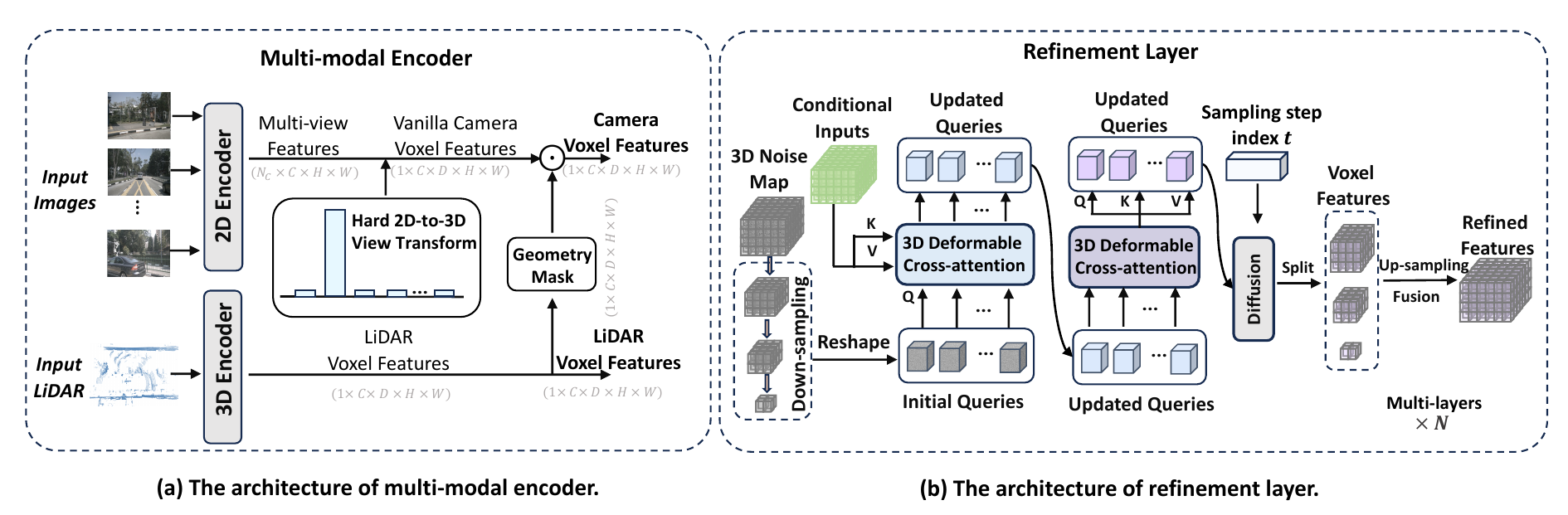}
\vspace{-2.0em}
\end{center}
   \caption{The detailed architectures of (a) multi-modal encoder and (b) refinement layer. The multi-modal encoder is a two-stream structure, comprising LiDAR and camera streams. The refinement layer consists of three main components, i.e., 3D deformable cross-attention, self-attention, and time diffusion modules.}
\label{Fig_detail}
\vspace{-1.5em}
\end{figure*}

% \vspace{-0.5em}

\subsection{OccGen Framework}
\label{overall}
We first depict the proposed ``noise-to-occupancy'' generative paradigm and then introduce the overall architecture. As shown in Fig.~\ref{Fig_method}, OccGen consists of a conditional encoder and a progressive refinement decoder.

% As illustrated in Figure \ref{Fig_method}, OccGen consists of two main components: the condition encoder and the progressive refinement module. The condition encoder is designed to process multi-modal inputs such as images and LiDAR point cloud data. The progressive refinement module aims to iteratively generate fine-grained occupancy predictions, using the multi-modal features as conditions.

% applies diffusion denoising, using the multi-modal features as conditions.

% \vspace{-0.5em}
\paragraph{Noise-to-occupancy generative paradigm.} We regard the 3D semantic occupancy prediction as a generative process, which progressively generates the surrounding 3D environment with detailed geometry and semantics from single or multi-modal inputs. The goal of noise-to-occupancy is to learn an occupancy perception model $f_{\theta}$ which can model the coarse-to-fine refinement of the dense 3D occupancy map through a total of $T$ diffusion steps:
\begin{equation}
Y_T \xrightarrow{f_{\theta}} Y_{T-1} \xrightarrow{f_{\theta}} \ldots \stackrel{f_{\theta}}{\rightarrow} Y_0,
\end{equation}
where the diffusion step $T \rightarrow 0$ represents the coarse-to-fine refinement process from a 3D Gaussian voxel map to the refined occupancy. Thus, the generative occupancy prediction paradigm can be formulated as:
\begin{equation}
\begin{aligned}
\Delta Y_{t} =f_{\theta}\left(x, t, Y_{t+1}\right), \quad 
Y_{t} =Y_{t+1} \oplus \Delta Y_t,
\end{aligned}
\end{equation}
where the model $f_{\theta}$ refines the current prediction occupancy by giving the diffusion step index $t$ and the previous-step prediction occupancy $Y_{t+1}$, and $\oplus$ is element-wise summation.

\paragraph{Conditional encoder.} 
\label{sec:encoder}
The conditional encoder consists of three main components: a multi-modal encoder, a fusion module, and an occupancy backbone. As shown in Fig.~\ref{Fig_detail} (a), the multi-modal encoder is a two-stream structure, comprising of LiDAR and camera streams. For the LiDAR stream, we follow VoxelNet \cite{Zhou2018VoxelNetEL} and 3D sparse convolutions \cite{Yan2018SECONDSE} to transform raw LiDAR points to LiDAR voxel features. In the camera stream, we utilize the pre-trained 2D backbones \cite{liu2021swin, Vaswani2017AttentionIA, he2016deep, dosovitskiy2020image} and Feature Pyramid Network (FPN) \cite{lin2017feature} to extract multi-view image features given multi-view images.
We obtain the vanilla camera voxel features through the 2D-to-3D view transformation. 

Different from the previous 2D-to-3D view transformation \cite{Philion2020LiftSS, li2022bevdepth, liu2022bevfusion, liang2022bevfusion} methods that estimate the probabilistic of a set of discrete depths, OccGen proposes a hard 2D-to-3D view transformation to guarantee more accurate depth. We opt for predicting a one-hot vector for depth, as opposed to utilizing softmax on discrete depth values when lifting each image individually into a frustum of features for each camera. However, obtaining one-hot encoding directly through $argmax$ operation is non-differentiable. To address this issue, we propose using Gumbel-Softmax \cite{jang2016categorical} to convert the predicted depth into one-hot encoding.

The previous multi-modal methods for 3D occupancy prediction do not pay much attention to the interaction between multi-modal features. Therefore, we propose a straightforward solution to fully exploit the geometry-aware correspondence between camera and LiDAR modalities. We directly generate a geometry mask by leveraging LiDAR voxel features and then applying it to the vanilla camera voxel features to get the camera voxel features. This feature aggregation strategy effectively bridges the gap between the camera voxel features and the true spatial distribution in the real-world scene. We follow \cite{wang2023openoccupancy} and fuse the camera and LiDAR voxel features using the adaptive fusion module:
\begin{equation}
\begin{aligned}
W & =\mathcal{G}_{\mathrm{C}}\left(\left[\mathcal{G}_{\mathrm{C}}\left(F_{p}\right), \mathcal{G}_{\mathrm{C}}\left(F_{c}\right)\right]\right), \\
F_m & =\sigma(W) \odot F_{p}+(1-\sigma(W)) \odot F_{c},
\end{aligned}
\end{equation}
where $\mathcal{G}_{\mathrm{C}}$ is the 3D convolution, $[\cdot, \cdot]$ is the concatenation along feature channel. $\sigma$ and $\odot$ denote the Sigmoid function and element-wise product, respectively. Finally, we fed the multi-modal voxel features $F_m$ into the occupancy backbone to get the multi-scale fusion features for the following progressive refinement decoder. Additional details and experiments of hard 2D-to-3D view transformation and geometry mask are presented in the supplementary materials.

\paragraph{Progressive Refinement Decoder.} The progressive refinement decoder of OccGen consists of a stack of refinement layers and an occupancy head. As illustrated in Fig.~\ref{Fig_detail} (b), the refinement layer takes as input the random noise map or the predicted noise map $Y_{t+1}$ from the last step, the current sampling step $t$, and the multi-scale fusion features $F_m$. The refinement layer utilizes efficient 3D deformable cross-attention and self-attention to refine the 3D Gaussian noise map. Compared with traditional deformable attention \cite{zhu2021deformable} in 2D vision, 3D deformable attention samples the interest points around the reference point in the 3D pixel coordinate system to compute the attention results. Mathematically, 3D deformable attention can be represented by the following general equation:
\begin{equation}
    \operatorname{DA}_{\operatorname{3D}}(q, p, F) = \sum_{k=1}^{N} A_{k} W_k F(p + \Delta p_{k}),
\end{equation}
where $q$ and $p$ denote the 3D query and 3D reference point, $F$ represents the flattened 3D voxel features, and $k$ indexes the sampled point from a total of $N$ points around the reference point $p$. $W_{k}$ represents the learnable weights for value generation, while $A_{k}$ corresponds to the learnable attention weight. $\Delta p_{k}$ denotes the predicted offset to the reference point $p
$, and $F(p+\Delta p_{k})$ signifies the feature at the location $p+\Delta p_{k}$ extracted through bilinear interpolation. For brevity, we present the formulation for single-head attention only. 

Directly operating on the original 3D Gaussian noise map $Y_t$ with high resolution is computationally intensive. Therefore, we first downsample it to obtain smaller multi-scale noise maps $Y_t^{i} \in \mathbb{R}^{\frac{D}{2^i} \times \frac{H}{2^i} \times \frac{W}{2^i} \times C_i}(i=1,2,3)$. Then, we reshape these downsampled multi-scale noise maps to obtain initial queries. For each initial query $q$ in the multi-scale noise maps $Y_t^{i}$, we get the corresponding reference points $p$ on conditional inputs based on their corresponding spatial and level positions. We get the updated queries using 3D deformable cross-attention ($\operatorname{DCA}_{\operatorname{3D}}$) by
\begin{equation}
\operatorname{DCA}_{\operatorname{3D}}\left(Y_t^i,F_m\right)= \sum_{n \in F_m} \operatorname{DA}_{\operatorname{3D}}\left( q, proj(q,n), F_m\right)
\end{equation}
where $n$ denotes the hit multi-scale features. For each query~$q$, we use $proj$ operation to obtain the reference point on multi-scale fusion features.

After one round of 3D deformable cross-attention, the initial queries gather knowledge from the condition inputs. To further enhance self-completion capability, we utilize the 3D deformable self-attention to update the queries,  
\begin{equation}
\operatorname{DSA}_{\operatorname{3D}}\left(Y_t^i, Y_t^i\right)= \sum_{n \in Y_t^i} \operatorname{DA}_{\operatorname{3D}}\left(q, p, \mathbf{Y}_t^i\right).
\end{equation}
Then, we split the learned queries into the down-sampled voxel sizes. We further apply a diffusion denoising step on the down-sampled multi-scale noise maps by
\begin{equation}
Y_t^{i} := \operatorname{Diff}(Y_t^{i}, \operatorname{ToEmbed}(t)),
\end{equation}
where $\operatorname{ToEmbed}(\cdot)$ denote the embedding network that transfors a step index $t$ from scalar into a feature vector. $\operatorname{Diff}(\cdot)$ represents the diffusion module that applies the scale and shift operation along the time embedding. Furthermore, we upsample and project the downsampled voxels to the size of the original 3D noise map and obtain the refined voxel features.
Finally, we obtain the 3D semantic occupancy by feeding the refined voxel features to the occupancy head. This process can be performed multiple times to progressively infer and refine the occupancy map by predicting and eliminating noise originating from a random Gaussian distribution.
\vspace{-1.0em}
\subsection{Training}
During training, we first construct a denoising diffusion process from the ground truth $Y_{0}$ to the 3D Gaussian noise map $Y_{T}$ and then train the progressive refinement decoder to reverse this process. Detailed information on the training procedure for OccGen is available in the supplementary materials.
% The training procedure for OccGen is provided in supplementary materials.
% \vspace{-0.5em}

\paragraph{Occupancy Corruption.} We add Gaussian noise to corrupt the encoded ground truth, obtaining the 3D Gaussian noise map $Y_{T}$. As shown in Eq.~\ref{eq:forward_ddpm}, the intensity of corruption noise is controlled by $\alpha_t$, which follows a monotonically decreasing schedule across different time steps $t \in [0, 1]$. Different noise scheduling strategies, including cosine schedule\cite{nichol2021improved} and linear schedule \cite{ho2020denoising}, are compared and discussed in the supplementary materials.
We found that the cosine schedule generally yields the best results in 3D semantic occupancy prediction.
% \vspace{-0.5em}

\paragraph{Loss Function.} The cross-entropy loss $\mathcal{L}_{\rm{ce}}$ and lovasz-softmax loss $\mathcal{L}_{\rm{ls}}$ \cite{berman2018lovasz} are widely used to optimize the networks for semantic segmentation tasks. Following \cite{wang2023openoccupancy, cao2022monoscene}, we also utilize affinity loss $\mathcal{L}_{\rm{scal}}^{\rm{geo}}$ and $\mathcal{L}_{\rm{scal}}^{\rm{sem}}$ to optimize the scene-wise and class-wise metrics (\ie, geometric IoU and semantic mIoU). Additionally, the depth loss $\mathcal{L}_{\rm{d}}$ \cite{li2022bevdepth} is used to optimize the predicted depth. Therefore, the overall loss function can be derived as:
\begin{equation}
\label{eq_loss}
    \mathcal{L}_{\rm{total}} = \mathcal{L}_{\rm{ce}} + \mathcal{L}_{\rm{ls}} + \mathcal{L}_{\rm{scal}}^{\rm{geo}} + \mathcal{L}_{\rm{scal}}^{\rm{sem}} + \mathcal{L}_{\rm{d}}.
\end{equation}

\vspace{-1.0em}
\subsection{Inference}
Given multi-scale fusion features as conditional inputs, OccGen samples a random noise map from a 3D Gaussian distribution and produces the occupancy in a coarse-to-fine manner. The inference procedure for OccGen is provided in supplementary materials. 
% \vspace{-0.5em}

\paragraph{Sampling Rule.} Following \cite{ji2023ddp}, we choose the DDIM strategy \cite{song2020denoising} for the sampling. In each sampling step $t$, the random noise map or the predicted noise map from the last step and the conditional multi-scale fusion features are sent to the progressive refinement decoder for occupancy prediction. After obtaining the predicted result of the current step, we compute the refined noise map for the next step using the reparameterization trick. Following \cite{chen2023diffusiondet, ji2023ddp}, we use the asymmetric time intervals (controlled by a hyper-parameter $td$) during the inference stage. We empirically set $td = 1$ in our method.
% \vspace{-0.5em}

\paragraph{Progressively Inference.}  According to the feature that the diffusion model can generate the distribution step by step, we can perform progressive inference to get fine-grained occupancy in a coarse-to-fine manner. Moreover, OccGen has a natural awareness of the prediction uncertainty. As a comparison, previous one-shot approaches for 3D semantic occupancy \cite{Zhang_2023_ICCV, li2023voxformer, wang2023openoccupancy, wei2023surroundocc} can only output a certain occupancy during the inference stage, and are unable to assess the reliability and uncertainty of model predictions.

\section{Experiments}
\label{sec:experiment}

\begin{table*}[ht]
	\setlength{\tabcolsep}{0.0035\linewidth}
	\newcommand{\classfreq}[1]{{~\tiny(\semkitfreq{#1}\%)}}  %
	\centering
    \caption{Semantic occupancy prediction results on nuScenes-Occupancy validation set. The $C,D,L,M$ denotes \textbf{camera, depth, LiDAR} and \textbf{multi-modal}. For \textbf{Surround=}\cmark, the method directly predicts surrounding semantic occupancy with 360-degree inputs. 
    % Otherwise, the method produces the results of each camera view and then concatenates them as surrounding outputs. 
    Best camera-only, LiDAR-only, and multi-modal results are marked \red{red}, \blue{blue}, and \textbf{black}, respectively.}
   \resizebox{1\linewidth}{!}{
	\begin{tabular}{l|c c | c c | c c c c c c c c c c c c c c c c}
 
		\toprule
		Method
		& \makecell[c]{Input}
		& \makecell[c]{Surround}
		& \makecell[c]{IoU}
            & \makecell[c]{mIoU}
		& \rotatebox{90}{\textcolor{barrier}{$\blacksquare$} barrier} 
		& \rotatebox{90}{\textcolor{bicycle}{$\blacksquare$} bicycle}
		& \rotatebox{90}{\textcolor{bus}{$\blacksquare$} bus} 
		& \rotatebox{90}{\textcolor{car}{$\blacksquare$} car} 
		& \rotatebox{90}{\textcolor{const. veh.}{$\blacksquare$} const. veh.} 
		& \rotatebox{90}{\textcolor{motorcycle}{$\blacksquare$} motorcycle} 
		& \rotatebox{90}{\textcolor{pedestrian}{$\blacksquare$} pedestrian} 
		& \rotatebox{90}{\textcolor{traffic cone}{$\blacksquare$} traffic cone} 
		& \rotatebox{90}{\textcolor{trailer}{$\blacksquare$} trailer} 
		& \rotatebox{90}{\textcolor{truck}{$\blacksquare$} truck} 
		& \rotatebox{90}{\textcolor{drive. suf.}{$\blacksquare$} drive. suf.} 
		& \rotatebox{90}{\textcolor{other flat}{$\blacksquare$} other flat} 
		& \rotatebox{90}{\textcolor{sidewalk}{$\blacksquare$} sidewalk} 
		& \rotatebox{90}{\textcolor{terrain}{$\blacksquare$} terrain} 
		& \rotatebox{90}{\textcolor{manmade}{$\blacksquare$} manmade} 
		& \rotatebox{90}{\textcolor{vegetation}{$\blacksquare$} vegetation} \\
		% & mIoU\\
		\midrule
		MonoScene~\cite{cao2022monoscene} & C & \xmark  & 18.4 & 6.9 & 7.1  & 3.9  &  9.3 &  7.2 & 5.6  & 3.0  &  5.9& 4.4& 4.9 & 4.2 & 14.9 & 6.3  & 7.9 & 7.4  & 10.0 & 7.6 \\
  
  		TPVFormer~\cite{huang2023tri} &C &  \cmark& 15.3 &  7.8 & 9.3  & 4.1  &  11.3 &  10.1 & 5.2  & 4.3  & 5.9 & 5.3&  6.8& 6.5 & 13.6 & 9.0  & 8.3 & 8.0  & 9.2 & 8.2 \\
    
            3DSketch~\cite{sketch} &  C\&D & \xmark& 25.6 & 10.7  & 12.0 &  5.1 &  10.7 &  12.4 & 6.5  & 4.0  & 5.0 & 6.3&  8.0&  7.2& 21.8 &  14.8 & 13.0 &  11.8 & 12.0 & 21.2 \\
            
            AICNet~\cite{aicnet} & C\&D   &  \xmark& 23.8 & 10.6  & 11.5  & 4.0  & 11.8  & 12.3&  5.1 & 3.8  & 6.2  & 6.0 & 8.2&  7.5&  24.1 & 13.0 & 12.8  & 11.5 & 11.6  &  20.2\\

            LMSCNet~\cite{lmscnet} & L &  \cmark& 27.3 & 11.5 & 12.4&  4.2 & 12.8  & 12.1  & 6.2  &  4.7 & 6.2 & 6.3&  8.8&  7.2& 24.2 & 12.3  & 16.6 & 14.1  & 13.9 & 22.2 \\

		  JS3C-Net~\cite{js3cnet} &L &  \cmark& 30.2  & 12.5 & 14.2 & 3.4  & 13.6  & 12.0  & 7.2  &       4.3 & 7.3 & 6.8&  9.2& 9.1 & 27.9 & 15.3  & 14.9 & 16.2  & 14.0 & \blue{24.9}  \\
		  \midrule
            C-OpenOccupancy \cite{wang2023openoccupancy} & C &  \cmark&19.3  & 10.3  &  9.9 & 6.8  & 11.2  & 11.5  & 6.3  & 8.4  & 8.6 & 4.3 & 4.2 & 9.9 & 22.0  & 15.8 & 14.1  & 13.5  & 7.3&10.2 \\
            L-OpenOccupancy \cite{wang2023openoccupancy} & L &  \cmark& 30.8  & 11.7 &  12.2  & 4.2  & 11.0  & 12.2  & 8.3  & 4.4  & 8.7 & 4.0& 8.4 & 10.3 & 23.5& 16.0 & 14.9 & 15.7  & 15.0 &17.9  \\
            OpenOccupancy \cite{wang2023openoccupancy} & M &   \cmark& 29.1 & 15.1 &  14.3  & 12.0  & 15.2  & 14.9  & 13.7  & 15.0  & 13.1 & 9.0 & 10.0 & 14.5 & 23.2 & 17.5 & 16.1  & 17.2 & 15.3  & 19.5  \\
		\midrule
            C-CONet \cite{wang2023openoccupancy}  & C &  \cmark&20.1  & 12.8&13.2  & 8.1 &  \red{15.4} &  17.2 & 6.3  & 11.2  & 10.0  &  8.3 & 4.7 & 12.1 & 31.4 & 18.8 & 18.7  & 16.3 & 4.8  &8.2  \\
            
            L-CONet \cite{wang2023openoccupancy} & L &  \cmark& 30.9  & 15.8 &  17.5  & \blue{5.2} & 13.3  & 18.1  & 7.8  & 5.4  & 9.6 & 5.6& 13.2 & 13.6 & 34.9 & 21.5  & 22.4 & 21.7  & 19.2 &23.5  \\

            CONet \cite{wang2023openoccupancy} & M &   \cmark& 29.5 & 20.1 &  23.3  & 13.3  & 21.2  & 24.3  & \textbf{15.3}  & 15.9  & 18.0 & 13.3 & 15.3 & 20.7 & 33.2 & 21.0 & 22.5  & 21.5 & 19.6  & 23.2  \\
            \midrule
            C-OccGen & C &   \cmark& \red{23.4}  & \red{14.5}  & \red{15.5} &  \red{9.1}  & 15.3  & \red{19.2}  & \red{7.3}  & \red{11.3} & \red{11.8}  & \red{8.9} & \red{5.9} & \red{13.7} &\red{34.8} & \red{22.0} & \red{21.8} & \red{19.5}  & \red{6.0} & \red{9.9} \\
            L-OccGen & L &   \cmark& \blue{31.6}  & \blue{16.8} &  \blue{18.8}  & 5.1  & \blue{14.8}  & \blue{19.6}  & \blue{7.0}  & \blue{7.7}  & \blue{11.5} & \blue{6.7} & \blue{13.9} &\blue{14.6} & \blue{36.4} & \blue{22.1} & \blue{22.8}  & \blue{22.3} & \blue{20.6} & 24.5   \\
            OccGen & M &   \cmark& \textbf{30.3} & \textbf{22.0} &  \textbf{24.9}  & \textbf{16.4}  & \textbf{22.5}  & \textbf{26.1}  & 14.0  & \textbf{20.1}  & \textbf{21.6} & \textbf{14.6} & \textbf{17.4} &\textbf{21.9} & \textbf{35.8} & \textbf{24.5} & \textbf{24.7}  & \textbf{24.0} & \textbf{20.5}  & \textbf{23.5}   \\
		\bottomrule
	\end{tabular}}\\
	\label{table:base_main}
\end{table*}

\subsection{Experimental Setup}
\paragraph{Dataset and Metrics.}
We evaluate our proposed OccGen on two benchmarks, i.e., nuScenes-Occupancy~\cite{wang2023openoccupancy} and SemanticKITTI~\cite{behley2019semantickitti}.
% The nuScenes dataset \cite{caesar2020nuscenes} is a public large-scale autonomous driving dataset that consists of 1000 driving scenes, which are captured by a variety of sensors (e.g. cameras, RADAR, LiDAR). 
% The nuScenes dataset contains 1,000 sequences. Each sequence lasts for around 20 seconds and the key-frames are annotated at 2Hz with 3D bounding boxes.  
The nuScenes-Occupancy extends the nuScenes~\cite{caesar2020nuscenes} to provide dense annotations on keyframes for 3D multi-modal semantic occupancy prediction. It covers 700 and 150 driving scenes in the training and validation sets of nuScenes. SemanticKITTI~\cite{behley2019semantickitti} contains 22 sequences including monocular images, LiDAR points, point cloud segentation labels and semantic scene completion annotations.
We follow previous works~\cite{wang2023openoccupancy, li2023voxformer, Zhang_2023_ICCV} to report the Intersection of Union (IoU) as the geometric metric and the mean Intersection over Union (mIoU) of each class as the semantic metric. Besides, the noise class is excluded from the evaluation.
% Besides, the noise class is ignored in the evaluation.

% \begin{figure}[t]
% \begin{center}
% %\fbox{\rule{0pt}{2in} \rule{0.9\linewidth}{0pt}}
% \includegraphics[width=0.8\linewidth]{sec/figures/inference.pdf}
% \vspace{-2.5em}
% \end{center}
%    \caption{Dynamic inference. The results of multiple inference on nuScenes-Occupancy.}
% \label{figintro}
% % \vspace{-2em}
% \end{figure}
\paragraph{Implementation Details.}
\label{networkdes}
We follow the same experiment settings of \cite{wang2023openoccupancy, Zhang_2023_ICCV} to make a fair comparison with previous methods \cite{cao2022monoscene, Zhang_2023_ICCV, li2023voxformer, wang2023openoccupancy} on both nuScenes-Occupancy and SemanticKITTI. 
% For the camera stream, we adopt ResNet-50 \cite{he2016deep} and FPN on multi-view images with the input image size of $1600 \time 900$ to obtain multi-scale camera features. In the LiDAR stream, we utilize VoxelNet \cite{Zhou2018VoxelNetEL} as the backbone and employ 3D sparse convolutions \cite{Yan2018SECONDSE} to obtain the LiDAR voxel features with 10 LiDAR sweeps as input. The point cloud is constrained within the range of $[\text{-}51.2m, 51.2m]$ for $X$ and $Y$ axis, and $[\text{-}5m, 3m]$ for the $Z$ axis. 
We simply stack six refinement layers with 3D deformable attention for the progressive refinement decoder. For training, we leverage the AdamW \cite{kingma2014adam} optimizer with a weight decay of 0.01 and an initial learning rate of 2e-4. We adopt the cosine learning rate scheduler with linear warming up in the first 500 iterations, and a similar augmentation strategy as BEVDet \cite{bevdet}. The models are trained for 24 epochs with a batch size of 8 on 8 V100 GPUs. The implementation details on nuScenes-Occupancy and SemanticKITTI are listed in supplementary materials.

\vspace{-0.5em}
\begin{table*}[ht]
    % \small
    % \setlength{\tabcolsep}{0.004\linewidth}
    \newcommand{\classfreq}[1]{{~\tiny(\semkitfreq{#1}\%)}}  %
    \newcommand{\clsname}[1]{\rotatebox{90}{\textcolor{#1}{$\blacksquare$} #1\classfreq{#1}}}
    \renewcommand{\tabcolsep}{2pt}
    \renewcommand\arraystretch{1.25}
    \centering
    \caption{\textbf{Semantic Scene Completion results on SemanticKITTI~\cite{behley2019semantickitti} validation set.} $^\dagger$ denotes the results provided by MonoScene~\cite{cao2022monoscene}.}
    \resizebox{1\linewidth}{!}{
    \begin{tabular}{l|c|ccccccccccccccccccc|c}
    \toprule
        Method & IoU
        & \clsname{road.}
        & \clsname{sidewalk.}
        & \clsname{parking.}
        & \clsname{otherground.}
        & \clsname{building.}
        & \clsname{car.}
        & \clsname{truck.}
        & \clsname{bicycle.}
        & \clsname{motorcycle.}
        & \clsname{othervehicle.}
        & \clsname{vegetation.}
        & \clsname{trunk.}
        & \clsname{terrain.}
        & \clsname{person.}
        & \clsname{bicyclist.}
        & \clsname{motorcyclist.}
        & \clsname{fence.}
        & \clsname{pole.}
        & \clsname{trafficsign.}
        & mIoU
        \\
    \midrule
        LMSCNet$^\dagger$ \cite{lmscnet} & 28.61 & 40.68 & 18.22 & 4.38 & 0.00 & 10.31 & 18.33 & 0.00 & 0.00 & 0.00 & 0.00 & 13.66 & 0.02 & 20.54 & 0.00 & 0.00 & 0.00 & 1.21 & 0.00 & 0.00 & 6.70 \\
        AICNet$^\dagger$ \cite{aicnet} & 29.59 & 43.55 & 20.55 & 11.97 & 0.07 & 12.94 & 14.71 & 4.53 & 0.00 & 0.00 & 0.00 & 15.37 & 2.90 & 28.71 & 0.00 & 0.00 & 0.00 & 2.52 & 0.06 & 0.00 & 8.31 \\
        JS3C-Net$^\dagger$ \cite{js3cnet} & 38.98 & 50.49 & 23.74 & 11.94 & 0.07 & 15.03 & 24.65 & 4.41 & 0.00 & 0.00 & 6.15 & 18.11 & 4.33 & 26.86 & 0.67 & 0.27 & 0.00 & 3.94 & 3.77 & 1.45 & 10.31 \\
        MonoScene \cite{cao2022monoscene} & 37.12 & 57.47 & 27.05 & 15.72 & \textbf{0.87} & 14.24 & 23.55 & 7.83 & 0.20 & 0.77 & 3.59 & 18.12 & 2.57 & 30.76 & 1.79 & 1.03 & 0.00 & 6.39 & 4.11 & 2.48 & 11.50 \\
        TPVFormer \cite{huang2023tri} & 35.61 & 56.50 & 25.87 & 20.60 & 0.85 & 13.88 & 23.81 & 8.08 & 0.36 & 0.05 & 4.35 & 16.92 & 2.26 & 30.38 & 0.51 & 0.89 & 0.00 & 5.94 & 3.14 & 1.52 & 11.36 \\
        VoxFormer \cite{li2023voxformer} & 44.02 & 54.76 & 26.35 & 15.50 & 0.70 & 17.65 & 25.79 & 5.63 & 0.59 & 0.51 & 3.77 & 24.39 & \textbf{5.08} & 29.96 & 1.78 & 3.32 & 0.00 & \textbf{7.64} & 7.11 & 4.18 & 12.35 \\
        OccFormer \cite{Zhang_2023_ICCV} & 36.50 & 58.85 & 26.88 & 19.61 & 0.31 & 14.40 & 25.09 & \textbf{25.53} & 0.81 & 1.19 & 8.52 & 19.63 & 3.93 & \textbf{32.62} & 2.78 & 2.82 & 0.00 & 5.61 & 4.26 & 2.86 & 13.46 \\ 
        Symphonize \cite{jiang2023symphonize} & 41.44 & 55.78 & 26.77 & 14.57 & 0.19 & \textbf{18.76} & \textbf{27.23} & 15.99 & \textbf{1.44} & 2.28 & 9.52 & \textbf{24.50} & 4.32 & 28.49 & 3.19 & \textbf{8.09} & 0.00 & 6.18 & \textbf{8.99} & \textbf{5.39} & 13.44 \\
        \hline
        \textbf{OccGen} (ours) & 36.87 & \textbf{61.28} & \textbf{28.30} & 20.42 & 0.43 & 14.49 & 26.83 & 15.49 & \textbf{1.60} & \textbf{2.53} & \textbf{12.83} & 20.04 & 3.94 & 32.44 & \textbf{3.20} & 3.37 & 0.00 & 6.94 & 4.11 & 2.77 & \textbf{13.74} \\ 
    \bottomrule
    \end{tabular}}
    \label{tab:semkitti_val}
\end{table*}
\vspace{-0.5em}

\subsection{Comparison with the state-of-the-art}
{\textbf{Results on nuScenes-Occupancy.}}
As shown in Tab.~\ref{table:base_main}, we report the quantitative comparison of existing LiDAR-based, camera-based, and multi-modal methods on nuScenes-Occupancy. 
% It is worth noting that L-OccGen solely employs the progressive refinement module. 
Observations show that OccGen outperforms all existing competitors, regardless of whether the camera-only, LiDAR-only, or multi-modal methods. Compared with the current SOTA method CO-Net \cite{wang2023openoccupancy}, OccGen achieves a remarkable boost of 1.7\%, 0.4\%, and 1.9\% mIoU for camera-only, LiDAR-only, and multi-modal benchmarks, respectively. This demonstrates the effectiveness of OccGen for semantic occupancy prediction. We also note that OccGen consistently delivers the best IoU results across almost all categories, which indicates that our method can better complete the scenes due to our coarse-to-fine generation property. It is also worth noting that OccGen with multi-modal inputs can improve camera-only and LiDAR-only by 7.5\% and 5.8\% mIoU, which demonstrates the effectiveness of the camera modality in capturing small objects (e.g., bicycle, pedestrian, motorcycle, traffic cone) and LiDAR modality on large objects structured regions (e.g., drivable surface, sidewalk, vegetation). This lays a solid foundation for us to further explore how to improve the role of images during fusion.

\noindent
{\textbf{Results on SemanticKITTI.}}
We also compare the proposed OccGen with the state-of-the-art vision-based works~\cite{Zhang_2023_ICCV, li2023voxformer, jiang2023symphonize} on SemanticKITTI. For a fair comparison, we removed the LiDAR stream and fusion module from the conditional encoder. As shown in Tab.~\ref{tab:semkitti_val}, we can see that OccGen achieves the highest mIoU compared with all existing competitors. Compared with the state-of-the-art OccFormer~\cite{Zhang_2023_ICCV}, our proposed method has an improvement of 0.3\% mIoU, demonstrating the effectiveness of OccGen for semantic scene completion. We also notice that the transformer-based methods~\cite{Zhang_2023_ICCV, Everingham2009ThePV, li2023voxformer, jiang2023symphonize} achieve higher performance than other previous methods. This reveals the superior capability of transformer-based structure in representation learning.

\subsection{Ablation Study}
% We conduct ablation studies for the proposed architecture, image backbone, input size, and denoising layers depth on the nuScenes-Occupancy. All models are trained with the same hyper-parameters for 24 epochs. 
\begin{table}[t]
   \scriptsize
   \scalebox{0.98}{
  	\begin{minipage}[b]{0.46\linewidth}
  		\centering
  		\makeatletter\def\@captype{table}
  		\setlength{\tabcolsep}{1mm}{
            \caption{Ablations on the conditional encoder and progressive refinement decoder on nuScenes-Occupancy under the multi-modal setting. (a), (b) and (c) denote our baseline, baseline ``with proposed encoder'' and ``with proposed decoder'', respectively.}
            \label{tab:ablation1}
    	  \begin{tabular}{l|c|c|c|c}
            %%%%%%%%%%%%%%%%%%%%%%%%%%%%%%%%%%%%%%%
            \toprule
            %%%%%%%%%%%%%%%%%%%%%%%%%%%%%%%%%%%%%%%%
            & Encoder \quad  & Decoder \quad & IoU  &  mIoU   \\
            \hline
            \noalign{\smallskip}
                (a) & - & - & 28.1 & 20.4\\ 
                (b) & \cmark & - & 28.6 & 20.7 \\ 
                (c) & - &\cmark  &30.1 & 21.6\\ 
                (d) & \cmark &  \cmark  &30.3 & 22.0\\ 
            \noalign{\smallskip}
            %%%%%%%%%%%%%%%%%%%%%%%%%%%%%%%%%%%%%%%%
            \bottomrule
            %%%%%%%%%%%%%%%%%%%%%%%%%%%%%%%%%%%%%%%%
            \end{tabular}}
  		\vspace{-8pt}
    \end{minipage}
    \hspace{8pt}
  	\begin{minipage}[b]{0.48\linewidth}
  		\centering
  		\makeatletter\def\@captype{table}
  		\setlength{\tabcolsep}{1mm}{
            \caption{Ablations on the progressive refinement decoder on nuScenes-Occupancy under the multi-modal setting. ``\textbf{DCA}'' and ``\textbf{DSA}'' denote the 3D deformable cross- and self-attention. }
            \label{tab:ablation2}
            \begin{tabular}{l|ccc}
            \toprule
            &Method & IoU & mIoU  \\
            \midrule
            \multirow{3}{*}{(a)} & w/o DSA & 30.1 & 21.4  \\
             & w/o CSA & 29.7 & 21.2\\
             & w/o CSA and DSA & 29.1 & 20.7\\
            \hline
            \multirow{2}{*}{(b)}  & DSA + DCA & 29.4 & 21.6  \\
             & CSA + DSA & 30.3 & 22.0\\
            \hline
            \multirow{2}{*}{(c)} & w/o Diffusion  & 29.3 & 21.7  \\
            & OccGen & 30.3 & 22.0  \\
            \bottomrule
            \end{tabular}
    	   }
  		
  		\vspace{-8pt}
    \end{minipage}}
\end{table}

\paragraph{Overall architecture.}
The ablation results on the conditional encoder and progressive refinement decoder are shown in Tab.~\ref{tab:ablation1}. It is obvious that both the conditional encoder and progressive refinement decoder can achieve performance improvement. We also notice that ``with proposed decoder'' has a higher performance than ``with proposed encoder'', demonstrating the effectiveness of our generative pipeline.

\paragraph{Progressive refinement decoder.}
We conduct the ablations on the detailed components of the progressive refinement decoder.
% The results are shown in Tab.~\ref{tab:ablation2} and Tab.~\ref{tab:ablation3}, respectively. 
From Tab.~\ref{tab:ablation2} (a) and (b), it is evident that both 3D deformable cross- and self-attention lead to noticeable improvements in results. Compared to self-attention, cross-attention has a greater impact on performance, which is intuitive: learning knowledge from conditional inputs is always more comprehensive. Additionally, we also observed that the order of DCA and DSA in the decoder has a certain impact on the results. We also see that removing the temporal diffusion process leads to a decrease in results from Tab.~\ref{tab:ablation2} (c).

\paragraph{Conditional encoder.}
We also conduct the ablations on the detailed components of the conditional encoder.  
From Tab.~\ref{tab:ablation3}, We also observe that the two solutions in the conditional encoder have both achieved promising performance. The reason is that the accurate depth estimation and geometry guidance can keep the fine-grained spatial structures. This effectively limits the impact of disruptive information from the images, leading to notable performance enhancements.

\vspace{-1.0em}
\subsection{Further Discussion}
\begin{figure*}[t]
\begin{center}
%\fbox{\rule{0pt}{2in} \rule{0.9\linewidth}{0pt}}
\includegraphics[width=0.87\linewidth]{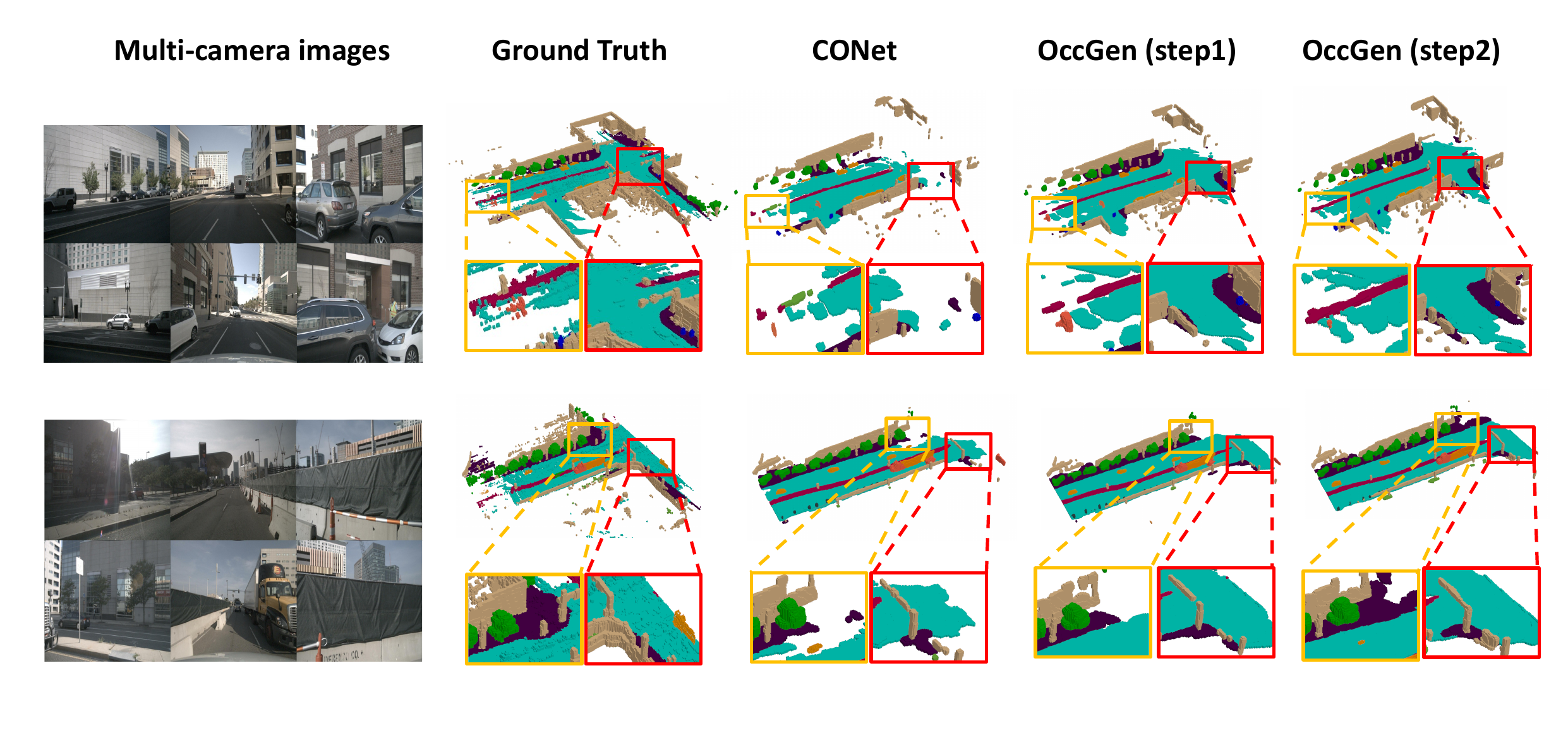}
\vspace{-2.5em}
\end{center}
   \caption{Qualitative results of the 3D semantic occupancy predictions on nuScenes-Occupancy. The leftmost column shows the input surrounding images, and the following four columns visualize the 3D semantic occupancy results from the ground truth, CONet\cite{wang2023openoccupancy}, OccGen(step1), and OccGen(step2). The regions highlighted by rectangles indicate that these areas have obvious differences (better viewed when zoomed in).}
\label{fig:compare}
\vspace{-0.8em}
\end{figure*}

% \begin{figure}[t]
%     \centering
%     \begin{subfigure}{0.49\linewidth}
%         \centering
%         \includegraphics[width=\linewidth]{figures/inference.pdf}
%         \caption{Training Results of BEVDet}
%     \end{subfigure}
%     \hfill
%     \begin{subfigure}{0.5\linewidth}
%         \centering
%         \includegraphics[width=\linewidth]{figures/uncer_vis.pdf}
%         \caption{Training Results of BEVFormer}
%     \end{subfigure}
%     \caption{Visualization of surrounding refinement capability. }
%     \label{Fig:ana}
%     \vspace{-8mm}
% \end{figure}
The desirable properties of OccGen compared with the previous discriminative occupancy methods in a single-forward process are shown in Fig.~\ref{fig:compare} and \ref{fig:uncer}. OccGen provides the flexibility to balance computational cost against prediction quality in a coarse-to-fine manner. Additionally, the stochastic sampling process enables the computation of voxel-wise uncertainty in the prediction.

% \vspace{-1.5em}

\paragraph{Progressive Refinement.} We evaluate OccGen with 1, 3, and 6 refinement layers by increasing their sampling steps from 1 to 10. The results are presented in Fig~\ref{fig:infer}. It can be seen that OccGen can continuously improve its performance by using more sampling steps. For instance, OccGen with 6 refinement layers shows an increase from 21.7\% mIoU (1 step) to 22.0\% mIoU (3 steps), and we visualize the inference results of different steps in Fig.~\ref{fig:compare}. In comparison to the previous single-step discriminative method, OccGen has the flexibility to balance computational cost against accuracy. This means our method can be adapted to different trade-offs between speed and accuracy under various scenarios without the need to retrain the network.

\paragraph{Efficiency vs. Accuracy.} We report the results of IoU and mIoU to represent the accuracy of different methods and latency(ms) to represent the efficiency of the models. The results are shown in Tab.~\ref{tab:latency}. Compared with the representative discriminative methods, OccGen achieves better results than state-of-the-art CONet~\cite{wang2023openoccupancy} when using only one sampling step, with comparable latency on the camera-only, and multi-modal settings. When adopting two sampling steps, the performance is further boosted to 21.8\% and 14.4\% on the multi-modal and camera-only benchmarks, at a loss of  $20\sim 50$ ms. These results show that OccGen can progressively refine the output occupancy multiple times with reasonable time cost. 

\begin{table}[t]
   \scriptsize
   \scalebox{0.98}{
  	\begin{minipage}[b]{0.43\linewidth}
  		\centering
  		\makeatletter\def\@captype{table}
  		\setlength{\tabcolsep}{1mm}{
            \caption{Ablations on the multi-modal encoder on nuScenes-Occupancy under the multi-modal setting.`\textbf{Hard LSS}'' and ``\textbf{Geo. Mask}'' denote hard 2D-to-3D view transformation and Geometry mask.}
            \label{tab:ablation3}
    	  \begin{tabular}{c|c|c|c}
            %%%%%%%%%%%%%%%%%%%%%%%%%%%%%%%%%%%%%%%
            \toprule
            %%%%%%%%%%%%%%%%%%%%%%%%%%%%%%%%%%%%%%%%
            Hard LSS \quad  & Geo. Mask \quad & IoU  &  mIoU   \\
            \hline
            \noalign{\smallskip}
                - & - & 29.8 & 21.4\\ 
                \cmark & - & 30.2 & 21.5\\ 
                - &\cmark  &30.3 & 21.6\\ 
                \cmark &  \cmark  &30.3 & 22.0\\ 
            \noalign{\smallskip}
            %%%%%%%%%%%%%%%%%%%%%%%%%%%%%%%%%%%%%%%%
            \bottomrule
            %%%%%%%%%%%%%%%%%%%%%%%%%%%%%%%%%%%%%%%%
            \end{tabular}
    	   }
  		
  		\vspace{-8pt}
    \end{minipage}
    \hspace{8pt}
  	\begin{minipage}[b]{0.56\linewidth}
  		\centering
  		\makeatletter\def\@captype{table}
  		\setlength{\tabcolsep}{1mm}{
            \caption{The latency, and performance on nuScenes-Occupancy under camera-only and multi-modal settings.}
            \label{tab:latency}
            \begin{tabular}{cccc}
            \toprule
            Models & Latency(ms) & IoU & mIoU \\
            \midrule
            C-Baseline~\cite{wang2023openoccupancy} & 172.4 & 19.3 & 10.3\\
            C-CONet~\cite{wang2023openoccupancy} & 285.7 & 20.1 & 12.8 \\
            C-OccGen(step1) & 294.1 & 23.0 & 14.2\\
            C-OccGen(step2) & 312.5 & 23.3 & 14.4\\
            \midrule
            Baseline~\cite{wang2023openoccupancy} & 243.9 & 29.1 & 15.1\\
            CONet~\cite{wang2023openoccupancy} & 344.8 & 29.5 & 20.1\\
            OccGen(step1) & 357.1 & 29.3 & 21.7\\
            OccGen(step2) & 400.0 & 29.7 & 21.8\\
            % OccGen(step3) & \textbf{88.7} & 117M &  \textbf{73.1} & \textbf{70.0}\\
            \bottomrule
        \end{tabular}
    	}
  		\vspace{-8pt}
    \end{minipage}}
\end{table}

\vspace{-1.0em}
\begin{figure}[t]
   \scriptsize
   \scalebox{0.95}{
  	\begin{minipage}[b]{0.49\linewidth}
  		\centering
  		\makeatletter\def\@captype{table}
  		\setlength{\tabcolsep}{1mm}{
            % \caption{Ablations on the multi-modal encoder on nuScenes-Occupancy under the multi-modal setting.`\textbf{Hard LSS}" and ``\textbf{Guidance}" denote hard 2D-to-3D view transformation and LiDAR guidance.}
            % \label{tab:ablation3}
            \centering
            \includegraphics[width=\linewidth]{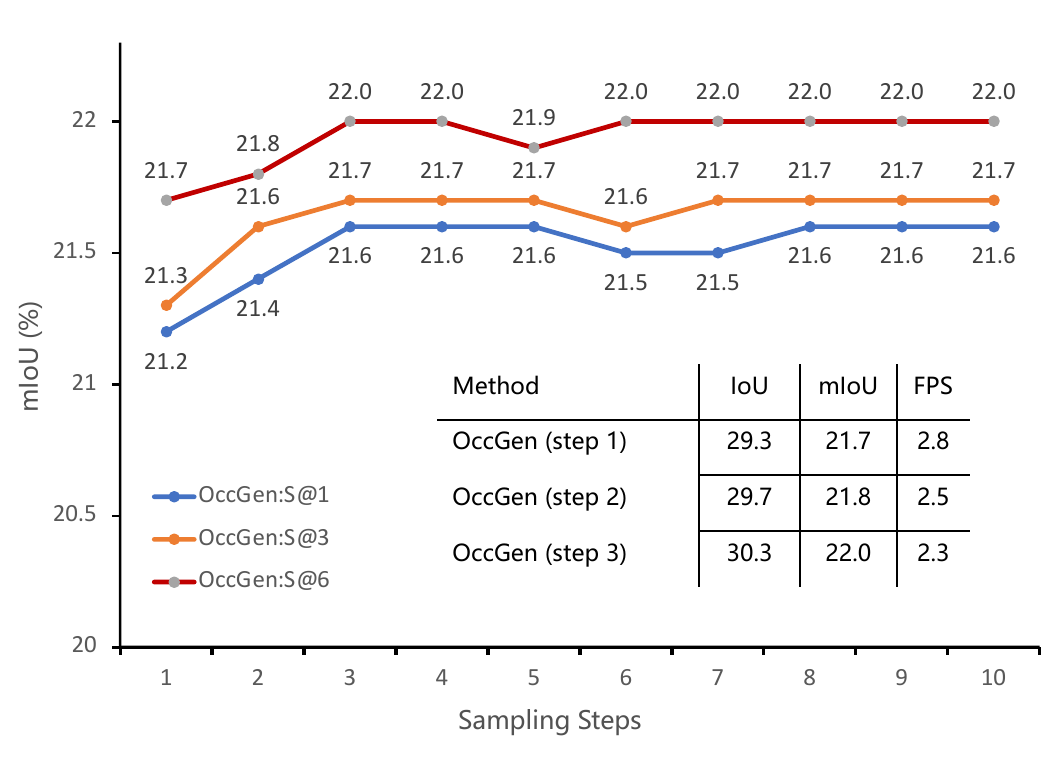}
            \caption{The results of multiple inferences on nuScenese-Occupancy under multi-modal setting.}
            \label{fig:infer}
    	  }
  		\vspace{-8pt}
    \end{minipage}
    \hspace{8pt}
  	\begin{minipage}[b]{0.5\linewidth}
  		\centering
  		\makeatletter\def\@captype{table}
  		\setlength{\tabcolsep}{1mm}{
            \centering
            \includegraphics[width=\linewidth]{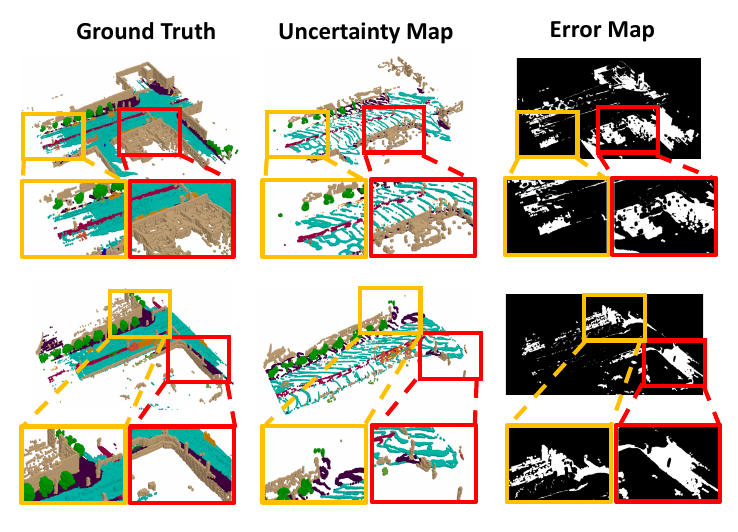}
            \caption{The visualization of uncertainty map and error map on nuScenese-Occupancy under multi-modal setting.}
            \label{fig:uncer}
    	}
  		\vspace{-8pt}
    \end{minipage}}
\end{figure}

% We note that DiffOcc consistently delivers the best IoU results across almost all categories in the third step, which indicates that our method can better complete the scenes due to our coarse-to-fine generation property. We also observe that camera-only methods are more time-consuming compared to LiDAR-only methods due to the 2D-to-3D view transformation. This indicates that a more efficient LSS method is urgent.
\vspace{1.0em}
\paragraph{Uncertainty Awareness.} In addition to the performance gains, the proposed OccGen can naturally provide uncertainty estimates. In the multi-step sampling process, we can simply count the voxels where the predicted result of each step differs from the result of the previous step, thereby obtaining an uncertainty occupancy result. We can see from Fig.~\ref{fig:uncer} that the areas with high uncertainty in the uncertainty map often align with those in the error map, which indicates incorrect prediction regions.
In comparison, OccGen offers a straightforward and inherently capable approach, whereas previous methods \cite{loquercio2020general, harakeh2020bayesod} require complicated modeling such as Bayesian networks.

\subsection{Qualitative Results}
\vspace{-0.1cm}

In Fig.~\ref{fig:compare}, we visualize the predicted results of 3D semantic occupancy on nuScenes-Occupancy from CONet \cite{wang2023openoccupancy} and our proposed OccGen. Compared with CONet, our method can better understand the scene-level semantic layout and perform local region completion. It is obvious that the regions of ``drivable surface'' and ``sidewalk'' predicted by our OccGen have higher continuity and integrity, and can effectively reduce a large number of hole areas compared with CONet. One more interesting observation is that due to the ground truth being initially constructed based on sparse LiDAR data, the shape of voxels in space is not very well-defined, especially in the drivable area. However, both CONet \cite{wang2023openoccupancy} and OccGen yields smoother predictions for these occupancy results.

\section{Conclusion}
In this paper, we propose OccGen, a simple yet powerful generative perception model for 3D semantic occupancy prediction. OccGen adapts a ``noise-to-occupancy'' generative paradigm, progressively inferring and refining the occupancy map from a random Gaussian distribution. OccGen consists of two main components: a conditional encoder that processes the multi-modal inputs as condition inputs and a progressive refinement decoder that produces fine-grained occupancy in a coarse-to-fine manner. OccGen has achieved state-of-the-art performance for 3D semantic occupancy prediction on nuScenes-Occupancy and SemanticKITTI. In addition, the proposed OccGen has shown desirable properties that discriminative models cannot achieve, such as progressive inference and uncertainty estimates. Currently, the latency of our OccGen is comparable to the previous state-of-the-art methods and has not achieved a significant speed advantage. Next, we will explore a more lightweight generative architecture for 3D semantic occupancy prediction.

\clearpage  % TODO REVIEW/FINAL: This \clearpage needs to be removed from both review and camera-ready versions.

% ---- Bibliography ----
%
% BibTeX users should specify bibliography style 'splncs04'.
% References will then be sorted and formatted in the correct style.
%
\bibliographystyle{splncs04}
\bibliography{main}

\clearpage

\begin{center}
\textbf{\large Supplementary Materials for \\ OccGen: Generative Multi-modal 3D Occupancy Prediction for Autonomous Driving}
\end{center}

In the supplementary material, we first present the methodology details of our proposed OcccGen, including hard 2D-to-3D view transformation, geometry mask, discriminative vs. generative modeling, and DDPM vs. DDIM. Then, we provide the details of datasets and implementation. Furthermore, we present additional experimental results to demonstrate the effectiveness of OccGen. Finally, we discuss the broader impact statement and limitations.

\section{Methodology Details}
\subsection{Hard 2D-to-3D View Transformation}
The previous LSS-based methods \cite{Philion2020LiftSS, liu2022bevfusion, liang2022bevfusion} associate a set of discrete depths for every pixel, covering the full range of potential depth values. These methods typically choose the softmax operation, which is a smooth approximation of $argmax$, which normalizes the vector, enabling gradient computation while the values can also represent probabilities.
Intuitively, this soft approach to depth prediction allows the network to learn depth information that is more conductive to feature optimization, rather than more precise depth information. As per the findings in \cite{li2022bevdepth}, this soft depth prediction fails to obtain precise depth information, consequently leading to the presence of ambiguous grids in the camera voxel. 
In light of this, we propose a hard 2D-to-3D view transformation method that utilizes hard Gumbel-Softmax~\cite{jang2016categorical} to obtain a deterministic discrete output vector. The key point lies in the fact that the $argmax$ operation for obtaining a one-hot vector is non-differentiable, making it impossible to calculate gradients, and consequently, network updates cannot be performed. Hard Gumbel-Softmax is a deterministic version of Gumbel-Softmax, where instead of sampling from the Gumbel-Softmax distribution. The formula for hard Gumbel-Softmax can be expressed as follows:
\begin{equation}
\bold{hard\_gumbel\_softmax} (z) = \bold{one\_hot}(\bold{argmax}((z + g)/\tau))
\end{equation}
where $z$ represents the logits. $g$ is sampled from the Gumbel distribution, typically calculated as $-log(-log(u))$ with u being a uniform random variable. $\tau$ is the temperature parameter controlling the softness of the distribution. When $\tau$ approaches zero, the hard Gumbel-Softmax approaches the one-hot encoding. The introduced hard 2D-to-3D view transformation enables gradient backpropagation during training and yields a definitive assignment of discrete depth during inference.
% we opt for predicting a one-hot vector for depth, as opposed to utilizing softmax on discrete depth values when lifting each image individually into a frustum of features for each camera, as shown in Fig~\ref{fig_intro}. The key point lies in the fact that the $argmax$ operation for obtaining a one-hot vector is non-differentiable, making it impossible to calculate gradients, and consequently, network updates cannot be performed. Therefore, we use Gumbel-Softmax \cite{jang2016categorical} to convert the predicted depth into one hot encoding $y_i$ as follows:
% \begin{equation}
% y_i=\frac{\exp \left(\left(\log \left(p_i\right)+G_i\right) / \tau\right)}{\sum_j \exp \left(\left(\log \left(p_j\right)+G_j\right) / \tau\right)},
% \end{equation}
% where $p_{i}$ represents the probability of discrete depth, $\tau$ is the temperature parameter controlling the ``softness" of the distribution and $G_i$ is a noise term.

\subsection{Geometry Mask}
The camera voxel features $F_c$ obtained through the hard 2D-to-3D view transformation module contain some voxels with misleading information, leading to a blurred feature distribution. This issue arises because all points along a camera ray in 3D space are projected to the same location on the 2D image plane, resulting in some voxels sharing the same image features. This inaccurate 3D spatial structure hinders subsequent feature fusion and adversely impacts the final detection results. To overcome this challenge, we propose a straightforward method to fully exploit the geometry-aware correspondence between camera and LiDAR modalities. We generate a geometry mask by leveraging the LiDAR voxel features and then applying it to the camera voxel features. This process effectively bridges the gap between the image voxel representation and the true spatial distribution, resulting in improved feature representation and better alignment with the real-world scene.

Due to the limitations of LiDAR sensors, the raw point cloud data only covers a portion of the real scene, resulting in incomplete object shapes. When the point cloud is regularized into voxel grids, only a small portion of these voxels contain non-empty information. As a result, the initial sparse voxels fail to adequately represent the 3D geometry-aware correspondence between LiDAR and camera features. Commonly used 3D sparse convolutions \cite{graham20183d, Yan2018SECONDSE} include two types, regular and submanifold sparse convolutions. The submanifold sparse convolution performs convolution operations only on the sparse features located at the center of its receptive field. Consequently, the output features of submanifold convolution maintain consistency with the positions of the input sparse features. On the other hand, regular sparse convolution performs convolution operations as long as there are features within its receptive field, resulting in the propagation of sparse features to their surrounding positions. This leads to a substantial increase in the density of the output sparse features. Through performing several 3D sparse convolutional operations, non-empty voxels are significantly increased, effectively covering more 3D space. Compared with the original LiDAR features, most of the 3D voxels are completed, and the generated 3D geometry-aware constraints can better reflect the real scene distribution.

Specifically, we can generate the geometry mask as follows,
\begin{equation}
    L_{mask} = f_{\text{dense}}(f_{\text{reg}}(f_{\text{sp}}(F_{\text{p}})))
\end{equation}
where $f_{sp}$ denotes a 3D sparse convolution block that mixes both regular and submanifold convolution, $f_\text{reg}$ denotes regular 3D sparse convolution, and $f_\text{dense}$ indicates that sparse camera voxel features are densified by padding zeros in empty positions.
However, there is no guarantee that all foreground objects can be adequately represented even after the expansion of sparse features. Directly utilizing the aforementioned constraints on the camera voxel grids can obtain numerous all-zero features, potentially losing meaningful features. Therefore, we employ a softmax operation along the height dimension to maintain the density of camera voxel features. Subsequently, we apply the 3D constraint to the image voxel feature, which can be represented as,
\begin{equation}
    {F}_{c} = \bold{Softmax}(L_{mask})\cdot F_{c}.
\end{equation}
This weight assignment reduces the influence of misleading or ambiguous features during the transformation, effectively guaranteeing the robustness of the spatial information.

% The camera voxel features $F_I$ learned from multi-view images contains the voxels with misleading information, leading to a blurred feature distribution. This issue arises because all points along a camera ray in 3D space are projected to the same location on the 2D image plane, resulting in some voxels sharing the same image features. This inaccurate 3D spatial structure hinders subsequent feature fusion and adversely impacts the final occupancy results. To overcome this challenge, we propose a straightforward solution to fully exploit the geometry-aware correspondence between camera and LiDAR modalities. We generate a 3D geometry-aware constraint $L_{const}$ by leveraging LiDAR voxel features and then applying it to the camera voxel features. This feature aggregation process effectively bridges the gap between the camera voxel features and the true spatial distribution in the real-world scene, which can be represented as:
% \begin{equation}
%     \hat{F}_{I} = \delta(L_{const})\cdot F_{I},
% \end{equation}
% where $\delta$ denotes the softmax operation along the channel dimension to maintain the density of camera voxel features to avoid numerous all-zero features.

\subsection{Discriminative vs. Generative Modeling}
\subsubsection{Discriminative Modeling.} Discriminative methods for 3D occupancy semantic prediction
aim to predict the occupancy and semantic labels of voxels in a 3D scene. These methods typically focus on learning the conditional distribution $P(Y|X)$, where $Y$ represents the occupancy and semantic labels of voxels, and $X$ represents the observed 3D scene data (e.g., point clouds or multi-view images). However, only learning these mapping between inputs and outputs may result in a limited understanding of the overall scene context. This can lead to incomplete or inaccurate scene completions, especially in complex scenes with intricate spatial relationships between objects. From the perspective of uncertainty, discriminative methods often do not explicitly model uncertainty in predictions, which can be crucial for 3D occupancy prediction where the inputs may be noisy or incomplete. This can lead to overconfident predictions in uncertain regions of the scene. Furthermore, these methods may struggle to incorporate prior knowledge or constraints into the learning process, which can be important for ensuring the coherence and consistency of the completed scene.

\subsubsection{Generative Modeling.} 
Generative modeling for 3D occupancy semantic prediction aim to generate complete 3D scenes from partial or incomplete input data. These methods often leverage generative models, such as GAN~\cite{goodfellow2014generative}, VAE~\cite{kingma2013auto} and diffusion model~\cite{ho2020denoising,song2020denoising}, to predict the semantic occupancy. Inspired by the great success of diffusion models, we adopt the diffusion model as our pipeline. In contrast to the previous discriminative works, which typically optimize the posterior probability $P(Y|X)$, our method establishes the joint probability $P(X, Y) = P(Y|X) * P(X)$. This optimization process can be denoted as 
\begin{equation}
\begin{split}
     \underset{f_{\theta}}{\mathrm{argmax}}\  p(X, Y) = \underset{f_{\theta}}{\mathrm{argmax}}\{
\underbrace{\log p(Y|X)}_{\textrm{data term}}
+\underbrace{\log p(X)}_{\textrm{prior term}}
\}.
\end{split}
\end{equation}
% \begin{equation}
% F_{*} = F()
% \mathrm{argmax}_{F}\, p(F|X_{i}) = {\underbrace{\log p(X_{i}|F)}_{\textrm{data term}}} +\underbrace{\log p(F)}_{\textrm{prior term}}.
% \end{equation}
The data term can be formally defined as the process of learning the mapping between multi-modal inputs and occupancy map,  while the prior term can be denoted as a denoising diffusion process for the multi-modal inputs. This holistic procedure can be seamlessly integrated into the framework of the conditional diffusion model. Generative models inherently capture uncertainty in the predictions. This can be useful in scenarios where the inputs are noisy or ambiguous, as the model can provide a measure of confidence in its predictions. In addition, generative models can generate more coherent and contextually relevant completions thanks to the prior knowledge about the scene incorporated in the training process.

% Discriminative methods for 3D occupancy semantic prediction
% aim to predict the occupancy and semantic lables of voxels in a 3D scene. These methods typically focus on learning the conditional distribution $P(Y|X)$, where $Y$ represents the occupancy and semantic lables of voxels, and $X$ represents the observed 3D scene data (e.g., point clouds or multi-view images). However, only learning these mapping between inputs and outputs may result in a limited understanding of the overall scene context. This can lead to incomplete or inaccurate scene completions, especially in complex scenes with intricate spatial relationships between objects. From the perspective of uncertainty, discriminative methods often do not explicitly model uncertainty in predictions, which can be cruial for 3D occupancy predictionn where the inputs may be noisy or incomplete. This can lead to overconfident predictions in uncertain regions of the scene. Furthermore, these methods may struggle to incorporate prior knowledge or constraints into the learning process, which can be important for ensuring the coherence and consistency of the completed scene.

\subsection{DDPM and DDIM}
A significant limitation of denoising diffusion probabilistic models (DDPM)~\cite{ho2020denoising} is their requirement for numerous iterations to generate high-quality samples. This is due to the generative process, which transforms noise into data, approximating the reverse of the forward diffusion process. The forward diffusion process may involve thousands of steps, necessitating iteration over all these steps to produce a single sample. This process is much slower compared to GANs, which only require one pass through a network.
To address this efficiency gap between DDPM and GAN, denoising diffusion implicit model (DDIM)~\cite{song2020denoising}. DDIM enables significantly faster sampling without compromising on the training objective, making generative models using this architecture competitive with GAN of the same model size and sample quality. This is achieved by estimating the cumulative effect of multiple Markov chain steps and incorporating them simultaneously. Since each Markov jump is modeled as a Gaussian distribution, they approximate the combined effect of multiple jumps by using a higher-variance Gaussian distribution with the same mean. It is worth noting that the sum of two Gaussians remains Gaussian. In this paper, we utilize DDPM and DDIM to corrupt the ground truth occupancy and progressively refine the noise map to obtain the final results. The results on DDIM and DDPM are listed in the experimental part. 

% \section{Additional Experiments}
\section{Dataset and Implementation}

\subsection{Details of Dataset.}
We provide results on nuScenes-Occupancy~\cite{wang2023openoccupancy}, Occ3D-nuScenes~\cite{tian2023occ3d} and SemanticKITTI~\cite{behley2019semantickitti}. nuScenes-Occupancy and Occ3D-nuScenes are extended from the large-scale nuScenes \cite{caesar2020nuscenes} dataset with dense semantic occupancy annotation. SemanticKITTI~\cite{behley2019semantickitti} provides dense semantic annotations for each LiDAR
sweep from the KITTI Odometry Benchmark~\cite{geiger2012we}. We will introduce nuScenes-Occupancy~\cite{wang2023openoccupancy}, Occ3D-nuScnes~\cite{tian2023occ3d} and SemanticKITTI~\cite{behley2019semantickitti} in sequence.

% \paragraph{nuScenes.} nuScenes \cite{caesar2020nuscenes} is a public large-scale autonomous driving dataset developed by the autonomous driving company nuTonomy. The dataset consists of 1000 driving scenes, with 700, 150, 150 for training, validation, and testing. Each driving scene is about 20s and the annotate frequency is 2Hz. The annotated frames are called samples, and the other frames are sweeps. The samples are manually annotated and have a set of annotations in the form of a 3D bounding box. The dataset is annotated with 3D bounding box, category information, and other attributes (visibility, state) of 23 object classes in total.  The data collection was mainly carried out in Boston and Singapore. The vehicle used for the collection is equipped with a 32-lane LiDAR at a capture frequency of 20Hz, and 6 cameras at a capture frequency of 12Hz. The resolution of the 6 cameras is $1600 \times 900$, and the cameras cover a 360-degree field of view, so they are very suitable for multi-modal fusion. 

\paragraph{nuScenes-Occupancy.} 
In the pursuit of establishing a large-scale surrounding occupancy perception dataset, wang \cite{wang2023openoccupancy} et al. introduced the nuScenes-Occupancy based on nuScenes~\cite{caesar2020nuscenes}. Notably, the nuScenes-Occupancy dataset boasts approximately 40 times more annotated scenes and about 5 times more annotated frames compared to the work presented in \cite{song2017semantic}. To efficiently achieve this extensive annotation and densification of occupancy labels, they introduced an Augmenting And Purifying (AAP) pipeline. The pipeline initiates annotation through the superimposition of multi-frame LiDAR points. Acknowledging the sparsity inherent in the initial annotation attributed to occlusion or limitations in LiDAR channels, they employed a strategy to augment it with pseudo-occupancy labels. These pseudo labels are constructed using a pre-trained baseline. To further enhance the quality of the annotations by reducing noise and artifacts, human efforts are enlisted in the purification process. 

\paragraph{Occ3D-nuScenes.} 
Occ3D-nuScenes~\cite{tian2023occ3d} is a comprehensive autonomous driving dataset comprising 700 training scenes and 150 validation scenes. Each frame in this dataset features a 32-beam LiDAR point cloud and six RGB images captured by six cameras positioned at different angles around the LiDAR. These frames are densely annotated with voxel-wise semantic occupancy labels. The dataset's occupancy scope spans from $-40m$ to $40m$ along the $X$ and $Y$ axes, and from $-1m$ to $5.4m$ along the $Z$ axis in the ego coordinate system. The voxel size for the occupancy labels is $0.4m \times 0.4m \times 0.4m$. Semantic labels in the dataset encompass 17 categories, which include 16 known object classes and an additional "empty" class.

\paragraph{SemanticKITTI.} 
The SemanticKITTI dataset~\cite{behley2019semantickitti} is focused on semantic scene understanding using LiDAR points and front cameras. OccGen is evaluated for semantic scene completion using the monocular left camera as input, following the approach of MonoScene~\cite{cao2022monoscene} and OccFormer~\cite{Zhang_2023_ICCV}. In this evaluation, the ground truth semantic occupancy is represented as $256 \times 256 \times 32$ voxel grids. Each voxel is $0.2m \times 0.2m \times 0.2m$ in size and is annotated with one of 21 semantic classes (19 semantics, 1 free, 1 unknown). Similar to previous work~\cite{cao2022monoscene, Zhang_2023_ICCV, li2023voxformer}, the dataset's 22 sequences are split into 10/1/11 for training/validation/testing.

\subsection{Implementation Details}
In the camera stream, we adopt the ResNet-50 \cite{he2016deep} model as our image backbone and employ the FPN \cite{lin2017feature} for multi-scale camera feature fusion, generating the image feature maps of size $6 \times 56 \times 100$, with 512 channels. Then, we utilize the proposed hard 2D-to-3D image view transformation to generate the camera voxel feature of size $128 \times 128 \times 10$, with 80 channels. In the LiDAR stream, the point cloud is constrained within the range of $[\text{-}51.2m, 51.2m]$for $X$ and $Y$ axis, and $[\text{-}5m, 3m]$ for the $Z$ axis. Voxelization is performed with a voxel size of $(0.1m, 0.1m, 0.1m)$. We utilize VoxelNet \cite{Zhou2018VoxelNetEL} as the backbone and employ FPN-3D \cite{lin2017feature} to produce the LiDAR voxel features of size $128 \times 128 \times 10$, with 80 channels. In order to fully exploit the implicit geometry-aware cues between camera and LiDAR modalities, we utilize the structure knowledge in LiDAR modality to guide the camera modality to learn geometry mask ${80 \times 128 \times 128 \times 10}$, which can improve the generalization capability of the fused voxel features significantly. Subsequently, we follow \cite{wang2023openoccupancy} and utilize ResNet3D and FPN-3D to generate multi-scale voxel features as condition input for the progressive refinement module. The progressive refinement consists of six refinement layers with 3D deformable attention. The refinement layer takes as input the random noise map or the predicted noise map from the last step, the current sampling step, and the multi-scale fusion features. We downsample the random noise map three times to obtain smaller multi-scale noise maps to avoid the high-resolution 3D Gaussian noise map. Then, we reshape these downsampled multi-scale noise maps to obtain initial queries. After the learning process of several refinement layers, we upsample and project the downsampled voxels to the size of the original 3D noise map and obtain the refined voxel features. Finally, we obtain the 3D semantic occupancy by feeding the refined voxel features to the occupancy head~\cite{wang2023openoccupancy} for full-scale evaluation. 
\begin{algorithm}[htbp]
  \caption{Training algorithm}
  \label{alg::training}
  \small
  \begin{algorithmic}[1]
    \Require
      Multi-modal inputs: $\{{X_{p}, X_{c}}\}$;
      GT occupancy: $Y$;
    \Ensure
      Training loss
    \State Extract multi-modal features $F_{p}$ and $F_{c}$. $F_{p}, F_{c}$  $\leftarrow$  $\bold{Extractor} $($X_{p}, X_{c}$)
    \State Aggregate the camera features with a geometry mask. $F_{c}$  $\leftarrow$  $\bold{Aggregate} $($F_{c}, M_{p}$)
    \State Obtain the multi-modal fusion features $F_{m}$. $F_{m}$ $\leftarrow$ $\bold{Fuser} (F_{p}, F_{c})$;
    \State Encoding the ground truth occupancy. $Y_{0}$ $\leftarrow$ $\bold{Encoding} (Y)$
    \State Construct noise signal and choose step index. $t$ $\leftarrow$ $\bold{Randint}(0, T)$, $\epsilon$ $\leftarrow$ $\bold{Randn}$(mean=0, std=1)
    \State Signal scaling. $Y_{0}$ $\leftarrow$ $\bold{Norm}(Y_{0})$
    \State Corrupt the occupancy input. $Y_{t}$ $\leftarrow$ $\bold{Schedule}(t) \times Y_{0} + (1 - \bold{Schedule}(t)) \times \epsilon$ 
    \State Obtain the downsampled multi-scale noise maps. $Y_{t}^{i}$ $\leftarrow$ $\bold{Downsample}(Y_{t})$
    \State Obtain the refined noise map. $Y_t$ $\leftarrow$ $\bold{Refine}(F_{m}, Y_{t+1}, t)$
    \State Predict the occupancy results. $\hat{Y_{t}}$ $\leftarrow$ $\bold{Voxel2Occ}(Y_{t})$
    \State Calculate the training loss $\mathcal{L}_{\rm{total}}$ (Eq.~\ref{eq_loss}).
  \end{algorithmic}
\end{algorithm}
% \vspace{-2.5em}

For the input, we follow the setting in \cite{wang2023openoccupancy} to take the image size as $900 \times 1600$, and utilize 10 sweeps to densify the LiDAR point cloud. During training, we adopt similar data augmentation strategies in \cite{wang2023openoccupancy} for both the image and LiDAR data. In our experiments, we utilize the AdamW \cite{adam} optimizer with a weight decay of 0.01 and an initial learning rate of $2e^{-4}$. We also utilize the cosine learning rate scheduler with linear warming up in the first 500 iterations. During training, we first construct the diffusion process from ground truth to noisy occupancy and then train the model to reverse this process. Algorithm \ref{alg::training} provides the pseudo-code of OccGen training procedure. The inference procedure of OccGen is a denoising sampling process from noise to 3D semantic occupancy. Starting from 3D voxel grids sampled in Gaussian distribution, the OccGen progressively refines its predictions, as shown in Algorithm \ref{alg::inference}. All models are trained and inferenced with a batch size of 8 on 8 V100 GPUs. 

\begin{algorithm}
  \caption{Inference algorithm}
  \label{alg::inference}
  \small
  \begin{algorithmic}[1]
    \Require
      Multi-modal inputs: $\{{X_{p}, X_{c}}\}$;
      Generative steps: $T$;
    \Ensure
      Prediction occupancy $\hat{Y}$
    \State Extract multi-modal features. $F_{p}, F_{c}$  $\leftarrow$  $\bold{Extractor} $($X_{p}, X_{c}$)
    \State Aggregate the camera features. $F_{c}$  $\leftarrow$  $\bold{Aggregate} $($F_{c}, M_{p}$)
    \State Obtain the multi-modal fusion features. $F_{m}$ $\leftarrow$ $\bold{Fuser} (F_{p}, F_{c})$;
    \State Initialize the noise map. $Y_{T}$ $\leftarrow$ $\bold{Randn}$(mean=0, std=1)
    \For{$i=1, 2, ..., T$} % For 语句，需要和EndFor对应
        \If{$i > 1$} % If 语句，需要和EndIf对应
            \State Update the current 3D noise map. $Y_{i}$ $\leftarrow$ $Y_{i-1}$ 
        \Else
            \State Obtain the refined noise map. $Y_{i}$ $\leftarrow$ $\bold{Refine}(F_{m}, Y_{i}, t)$ 
            \State Predict the occupancy results. $\hat{Y_{i}}$ $\leftarrow$ $\bold{Voxel2Occ}(Y_{i})$
            \State Obtain the noise map for the next evaluation. $Y_{i}$ $\leftarrow$ $\bold{DDIM}(Y_{i}, i)$ 
        \EndIf
    \EndFor
  \end{algorithmic}
\end{algorithm}

\section{Additional Experiments}
% \paragraph{Results on nuScenes-Occupancy.}

\paragraph{Results on Occ3D-nuScenes.}
We also compare our OccGen with the state-of-the-art vision-based 3D occupancy prediction methods~\cite{Zhang_2023_ICCV, wei2023surroundocc, li2023voxformer, li2023fb} on Occ3D-nuScenes~\cite{tian2023occ3d}. For a fair comparison, we removed the LiDAR stream and fusion module from the conditional encoder and followed the same backbone and image size of FB-Occ~\cite{li2023fb}. As shown in Tab.~\ref{tab:occ3d}, we can see that OccGen achieves the highest mIoU compared with all existing SOTA methods, demonstrating the effectiveness of OccGen for semantic scene completion. 
\begin{table}[ht]
\begin{center}
\vspace{-1.5em}
\caption{Semantic occupancy prediction results on Occ3D-nuScenes validation set.}
\label{tab:occ3d}
\resizebox{0.55\textwidth}{!}{
\begin{tabular}{ccccc}
\toprule
Method & Backbone & Image Size & mIoU \\
\midrule
OccFormer~\cite{Zhang_2023_ICCV} & ResNet-50 & 900 $\times$ 1600 & 36.5 \\
SurroundOcc~\cite{wei2023surroundocc} & InternImage-B & 900 $\times$ 1600 &  40.7  \\
VoxFormer~\cite{li2023voxformer} & ResNet-101 & 900 $\times$ 1600 &  40.7 \\
FB-Occ~\cite{li2023fb} & ResNet-50 & 900 $\times$ 1600 & 41.8 \\
Ours & ResNet-50 & 900 $\times$ 1600 & {\bfseries 42.6} \\
\bottomrule
\end{tabular}}
\end{center}
\vspace{-1.5em}
\end{table}

\paragraph{Additional Results on nuScenes-Occupancy.} We report the results of IoU and mIoU to represent the accuracy of different methods, and Params and FPS to represent the efficiency of the models. The results are shown in Tab.~\ref{table:effciency}. Compared with the representative discriminative methods, OccGen achieves better results when using only one sampling step, with fewer parameters and comparable FPS on the camera-only, LiDAR-only, or multi-modal methods. When adopting three sampling steps, the performance is further boosted to 22.0\%, 16.8\%, and 14.5\% on the multi-modal, camera-only, and LiDAR-only benchmarks, at a loss of  $0.3 \sim 0.5$ FPS. These results show that OccGen can progressively refine the output occupancy multiple times with reasonable time cost. We note that OccGen consistently delivers the best IoU results across almost all categories in the third step, which indicates that our method can better complete the scenes due to our coarse-to-fine generation property. We also observe that camera-only methods are more time-consuming compared to LiDAR-only methods due to the 2D-to-3D view transformation. This indicates that a more efficient LSS method is urgent.

\begin{table*}[t]
	\setlength{\tabcolsep}{0.0035\linewidth}
	\newcommand{\classfreq}[1]{{~\tiny(\semkitfreq{#1}\%)}}  %
	\centering
    \caption{Performance on nuScenes-Occupancy (validation set). We report the geometric metric IoU, semantic metric mIoU, IoU, parameters and FPS for each semantic class. The $C,L,M$ denotes \textit{camera, LiDAR} and \textit{multi-modal}. The best results are {in boldface}	(Best camera-only, LiDAR-only, and multi-modal results are marked \red{red}, \blue{blue}, and \textbf{black}, respectively.}
    \label{table:effciency}
   \resizebox{1\linewidth}{!}{
	\begin{tabular}{l|c | c c c c | c c c c c c c c c c c c c c c c}
 
		\toprule
		Method
		& \makecell[c]{Input}
		& \makecell[c]{IoU}
            & \makecell[c]{mIoU}
            & \makecell[c]{Params}
            & \makecell[c]{FPS}
		& \rotatebox{90}{\textcolor{barrier}{$\blacksquare$} barrier} 
		& \rotatebox{90}{\textcolor{bicycle}{$\blacksquare$} bicycle}
		& \rotatebox{90}{\textcolor{bus}{$\blacksquare$} bus} 
		& \rotatebox{90}{\textcolor{car}{$\blacksquare$} car} 
		& \rotatebox{90}{\textcolor{const. veh.}{$\blacksquare$} const. veh.} 
		& \rotatebox{90}{\textcolor{motorcycle}{$\blacksquare$} motorcycle} 
		& \rotatebox{90}{\textcolor{pedestrian}{$\blacksquare$} pedestrian} 
		& \rotatebox{90}{\textcolor{traffic cone}{$\blacksquare$} traffic cone} 
		& \rotatebox{90}{\textcolor{trailer}{$\blacksquare$} trailer} 
		& \rotatebox{90}{\textcolor{truck}{$\blacksquare$} truck} 
		& \rotatebox{90}{\textcolor{drive. suf.}{$\blacksquare$} drive. suf.} 
		& \rotatebox{90}{\textcolor{other flat}{$\blacksquare$} other flat} 
		& \rotatebox{90}{\textcolor{sidewalk}{$\blacksquare$} sidewalk} 
		& \rotatebox{90}{\textcolor{terrain}{$\blacksquare$} terrain} 
		& \rotatebox{90}{\textcolor{manmade}{$\blacksquare$} manmade} 
		& \rotatebox{90}{\textcolor{vegetation}{$\blacksquare$} vegetation} \\
		% & mIoU\\
		  \midrule
            C-Baseline \cite{wang2023openoccupancy} & C &19.3  & 10.3 & 93M & 5.8 &  9.9 & 6.8  & 11.2  & 11.5  & 6.3  & 8.4  & 8.6 & 4.3 & 4.2 & 9.9 & 22.0  & 15.8 & 14.1  & 13.5  & 7.3&10.2 \\
            
            L-Baseline \cite{wang2023openoccupancy} & L & 30.8  & 11.7 & 63M & 6.9 &  12.2  & 4.2  & 11.0  & 12.2  & 8.3  & 4.4  & 8.7 & 4.0& 8.4 & 10.3 & 23.5& 16.0 & 14.9 & 15.7  & 15.0 &17.9  \\

            M-Baseline \cite{wang2023openoccupancy} & M & 29.1 & 15.1 & 117M & 4.1 & 14.3  & 12.0  & 15.2  & 14.9  & 13.7  & 15.0  & 13.1 & 9.0 & 10.0 & 14.5 & 23.2 & 17.5 & 16.1  & 17.2 & 15.3  & 19.5  \\
		\midrule
            C-CONet \cite{wang2023openoccupancy}  & C &20.1  & 12.8 & 111M & 3.5  & 13.2  & 8.1 &  \red{15.4} &  17.2 & 6.3  & 11.2  & 10.0  &  8.3 & 4.7 & 12.1 & 31.4 & 18.8 & 18.7  & 16.3 & 4.8  &8.2  \\
            
            L-CONet \cite{wang2023openoccupancy} & L & 30.9  & 15.8 & 63M & 4.0  &  17.5  & \blue{5.2} & 13.3  & 18.1  & 7.8  & 5.4  & 9.6 & 5.6& 13.2 & 13.6 & 34.9 & 21.5  & 22.4 & 21.7  & 19.2 &23.5  \\

            M-CONet \cite{wang2023openoccupancy} & M & 29.5 & 20.1 & 137M & 2.9  &  23.3  & 13.3  & 21.2  & 24.3  & \textbf{15.3}  & 15.9  & 18.0 & 13.3 & 15.3 & 20.7 & 33.2 & 21.0 & 22.5  & 21.5 & 19.6  & 23.2  \\
            \midrule
            C-OccGen (step1) & C & 23.0& 14.2 & 110M & 3.4  & 15.5 & 9.1 & 15.0 & 18.9 & 6.6 & 11.6 & 11.4 & 8.8 & 5.4 & 13.1 & 34.4 & 21.4 & 21.6 & 18.8 & 5.6 & 9.6 \\
            C-OccGen (step2) & C & 23.3 & 14.4 & 110M & 3.2 & 14.8 & 8.5 & 15.2 & 19.0 & 7.3 & 11.4 & 11.9 & 8.3 & 6.0 & 13.9 & 34.6 & 22.0 & 21.6 & 19.5 & 5.7 & 9.8 \\
            C-OccGen (step3) & C & \red{23.4}  & \red{14.5} & 110M & 3.0  & \red{15.5} &  \red{9.1}  & 15.3  & \red{19.2}  & \red{7.3}  & \red{11.3} & \red{11.8}  & \red{8.9} & \red{5.9} & \red{13.7} &\red{34.8} & \red{22.0} & \red{21.8} & \red{19.5}  & \red{6.0} & \red{9.9} \\
            \midrule
            L-OccGen (step1) & L & 31.1  & 16.1 & 62M & 4.0 & 17.6 & 4.1 & 14.3 & 19.1 & 6.6 & 7.1 & 11.0 & 6.2 & 13.2 & 14.3 & 35.8 & 21.3 & 22.2 & 20.9 & 20.1 & 24.2  \\
            L-OccGen (step2) & L & 31.4  & 16.6 & 62M & 3.9 &  18.7 & 5.1 & 15.0 & 19.3 & 7.3 & 7.8 & 11.2 & 6.3 & 13.7 & 14.3 & 36.3 & 21.9 & 22.7 & 21.9 & 20.2 & 24.1 \\
            L-OccGen (step3) & L & \blue{31.6}  & \blue{16.8} & 62M & 3.7 &  \blue{18.8}  & 5.1  & \blue{14.8}  & \blue{19.6}  & \blue{7.0}  & \blue{7.7}  & \blue{11.5} & \blue{6.7} & \blue{13.9} &\blue{14.6} & \blue{36.4} & \blue{22.1} & \blue{22.8}  & \blue{22.3} & \blue{20.6} & 24.5   \\
            \midrule
            OccGen (step1) & M & 29.3 & 21.7 & 117M & 2.8  &  25.4 & 16.6 & 22.2 & 26.0 & 13.4 & 19.9 & 21.8 & 14.6 & 17.3 & 22.1 & 35.4 & 24.1 & 24.1 & 22.8 & 19.5 & 22.3 \\
            OccGen (step2) & M & 29.7 & 21.8 & 117M & 2.5  &  24.8 & 16.8 & 22.4 & 25.9 & 13.8 & 20.3 & 21.7 & 14.6 & 17.5 & 21.9 & 35.2 & 24.5 & 24.3 & 23.5 & 19.5 & 22.5   \\
            OccGen (step3) & M & \textbf{30.3} & \textbf{22.0} & 137M & 2.3  &  \textbf{24.9}  & \textbf{16.4}  & \textbf{22.5}  & \textbf{26.1}  & 14.0  & \textbf{20.1}  & \textbf{21.6} & \textbf{14.6} & \textbf{17.4} &\textbf{21.9} & \textbf{35.8} & \textbf{24.5} & \textbf{24.7}  & \textbf{24.0} & \textbf{20.5}  & \textbf{23.5}   \\
		\bottomrule
	\end{tabular}}\\
\end{table*}

\paragraph{Scaling factor.} The performance of different scaling factors is shown in Tab.~\ref{tab:sf}. As can be seen, we found the lower scaling factor 0.001 has achieved a bit lower performance than 0.01. A larger scaling factor means more noise is added to the estimate, typically resulting in faster convergence but potentially reducing the quality of sampling. Conversely, a smaller scaling factor reduces the noise level but may require more iteration steps to converge, thereby increasing sampling time.

\paragraph{Noise schedule.} As shown in Tab.~\ref{tab:ns}, we compare the
effectiveness of the cosine schedule \cite{nichol2021improved} and linear schedule \cite{ho2020denoising} in OccGen for occupancy prediction. We observe that the model using a cosine schedule achieves better performance (22.0\% vs. 21.4\%). The possible reason is that the cosine schedule allows for a smooth reduction in noise, promoting more stable learning dynamics and the linear schedule may sometimes exhibit a more abrupt transition, and its impact on model convergence and sample quality can differ from the cosine schedule.

\paragraph{Sampling strategy.} As shown in Tab.~\ref{tab:ss}, we compare the effectiveness of the DDIM \cite{song2020denoising} and DDPM \cite{ho2020denoising} sampling strategies in OccGen, and find that the model using DDIM is better than DDPM. DDIM uses a non-Markovian diffusion process to accelerate sampling and DDPM is defined as the reverse of a Markovian diffusion process. 

\begin{table}[t]
    \centering
    \small
    \caption{The diffusion settings of progressive refinement layer on nuScenes-Occupancy. We report Iou and mIoU. Default settings are marked in \colorbox{bestcolor}{gray}.}
    \vspace{-1.5em}
    \label{tab:ds}
    \begin{subtable}{0.28\textwidth}
        \centering
        \caption{\textbf{Scaling factor}. The best scaling factor is 0.01.}
        \label{tab:sf}
        \scalebox{0.8}{
        \begin{tabular}{ccc}
        \toprule
        scale \quad & IoU & mIoU \\
        \hline
        0.001 \quad & 30.0  & 21.8  \\
        \bestcell{0.01} \quad & \bestcell{30.3} & \bestcell{22.0} \\
        % 0.1 & 5.1 & 5.5  \\
        \bottomrule
        \end{tabular}
        }
    \end{subtable}
    \hfill
    \begin{subtable}{0.28\textwidth}
        \centering
        \caption{\textbf{Noise schedule.} Consine works best. }
        \label{tab:ns}
        \scalebox{0.8}{
        \begin{tabular}{ccc}
        \toprule
        type \quad & IoU & mIoU\\
        \hline
        \bestcell{cosine} \quad  & \bestcell{30.3} & \bestcell{22.0} \\
        linear \quad & 29.9 & 21.4 \\
        \bottomrule
        \end{tabular}
        }
    \end{subtable}
    \hfill
    \begin{subtable}{0.35\textwidth}
        \centering
        \caption{\textbf{Sampling strategy.}  Using DDIM works best. }
        \label{tab:ss}
        \scalebox{0.8}{
        \begin{tabular}{ccc}
        \toprule
        type \quad  &  IoU &  mIoU \\
        \hline
        \bestcell{DDIM} \quad  & \bestcell{30.3} & \bestcell{22.0} \\
        DDPM \quad & 29.2 & 21.7 \\
        \bottomrule
        \end{tabular}
        }
    \end{subtable}
    \vspace{-1.0em}
    \hfill
    % \begin{subtable}{0.25\textwidth}
    %     \centering
    %     \scalebox{0.92}{
    %     \begin{tabular}{x{24}x{24}x{24}x{24}x{24}}
    %     \shline
    %     type & IoU & mIoU \\
    %     \shline
    %     \bestcell{3D Conv} & \bestcell{30.3} & \bestcell{22.0} \\
    %     3D Attn. & 29.2 & 21.7 \\
    %     \shline
    %     \end{tabular}
    %     }
    %     \caption{Ablations on the progressive refinement decoder in OccGen. }
    %     \label{tab:2d}
    % \end{subtable}
\end{table}

\begin{table}[t]
  \centering
  \normalsize
  \caption{Ablations on hard 2D-to-3D view transformation in OccGen under the multi-modal setting. ``\textit{Hard LSS}" and ``\textit{Depth Supervision}" denote hard 2D-to-3D view transformation and the generated depth ground truth following \cite{li2022bevdepth, wang2023openoccupancy} , respectively.} 
  \resizebox{0.42\textwidth}{!}{
    \begin{tabular}{c|c|c|c|c}
        %%%%%%%%%%%%%%%%%%%%%%%%%%%%%%%%%%%%%%%
         \toprule
        %%%%%%%%%%%%%%%%%%%%%%%%%%%%%%%%%%%%%%%%
           &  Hard LSS & Depth Supervision &  IoU & mIoU\\
        \hline\hline
       \noalign{\smallskip}
          (a) & - & - & 25.1 & 19.4\\ 
          (b) & \cmark & - & 28.6 & 20.6\\ 
          (c) & - &\cmark  &29.5 & 20.1\\ 
          (d) & \cmark &  \cmark  &29.4 & 20.8\\ 
         \noalign{\smallskip}
        %%%%%%%%%%%%%%%%%%%%%%%%%%%%%%%%%%%%%%%%
         \bottomrule
        %%%%%%%%%%%%%%%%%%%%%%%%%%%%%%%%%%%%%%%%
    \end{tabular}}
    \label{tab:abl_depth} % end caption
    \vspace{-1.0em}
\end{table}

\paragraph{The effectiveness of hard 2D-to-3D view transformation.} We also conduct experiments to fully exploit the effectiveness of hard 2D-to-3D view transformation under the multi-modal setting. The results are shown in Tab.~\ref{tab:abl_depth} We observe that ``Depth supervision'' proposed in BEVDepth \cite{li2022bevdepth} can boost the performance of occupancy prediction significantly. This indicates that the accurately predicted depth can lead to more complete occupancy. We also note that our proposed hard 2D-to-3D view transformation can achieve comparable results without adopting depth supervision, which demonstrates the effectiveness of the hard Gumbel softmax on depth prediction.

\paragraph{Different Experiment Setting.}
In this subsection, we ablate the different experiment settings (\eg, input size, backbone selection, number of refinement layers) in Tab.~\ref{tab:abl_sett}. For the camera-based OccGen, using a larger input size ($1600\times 900$) relatively improves IoU and mIoU by 7.3\% and 11.5\%. Besides, replacing ResNet-50 with ResNet-101 can further improve the performance of mIoU. For the LiDAR-based OccGen, it is observed that utilizing multi-sweeps as input (following \cite{Yin2020Centerbased3O, Yan2018SECONDSE, Lang2019PointPillarsFE}, 10 sweeps are used) perform well the single-sweep counterpart on IoU and mIoU. For the multi-modal OccGen, we observe that the number of refinement layers has a discernible impact on the performance. The performance tends to increase with a greater number of layers.

\begin{table}[t]
  \centering
  \setlength{\tabcolsep}{1.8mm} 
  \small
    \caption{Ablation study of backbone selection, input size of different modality, and number of denoising layers. \textit{C,L} denotes camera and LiDAR.} 
  \resizebox{0.65\textwidth}{!}{
    \begin{tabular}{l|c|c|c|c|c|c}
        %%%%%%%%%%%%%%%%%%%%%%%%%%%%%%%%%%%%%%%
         \toprule
        %%%%%%%%%%%%%%%%%%%%%%%%%%%%%%%%%%%%%%%%
          & Method &  2D Backbone &  Input Size &  Layers & IoU & mIoU\\
        \hline\hline
       \noalign{\smallskip}
          \multirow{3}{*}{(a)} \quad & C-OccGen & R-50 & $704\times 256$ & six &21.8 &13.0 \\ 
          & C-OccGen & R-50 & $1600\times 900$ & six & 23.4 & 14.5\\ 
          & C-OccGen & R-101 & $1600\times 900$ & six & 23.3 & 15.0\\ 
        \noalign{\smallskip}
        \hline
        \noalign{\smallskip}
          \multirow{2}{*}{(b)} \quad & L-OccGen & - & 1 sweep & six & 30.4 & 15.9\\ 
          & L-OccGen & - & 10 sweeps & six & 31.6 & 16.2\\ 
        \noalign{\smallskip}
        \hline
        \noalign{\smallskip}
          \multirow{5}{*}{(c)} \quad & OccGen &  R-50  & \makecell[c]{$1600\times 900$ \\10 sweeps} & one & 29.4& 21.6\\ 
          
          & OccGen &  R-50  & \makecell[c]{$1600\times 900$ \\10 sweeps} & three & 29.9& 21.7\\ 
          
          & OccGen &  R-50  & \makecell[c]{$1600\times 900$ \\10 sweeps} & six & 30.4 & 22.0 \\ 
          
         \noalign{\smallskip}
        %%%%%%%%%%%%%%%%%%%%%%%%%%%%%%%%%%%%%%%%
         \bottomrule
        %%%%%%%%%%%%%%%%%%%%%%%%%%%%%%%%%%%%%%%%
    \end{tabular}}
    \label{tab:abl_sett} % end caption
    \vspace{-0.5em}
\end{table}

\begin{figure}
\begin{center}
%\fbox{\rule{0pt}{2in} \rule{0.9\linewidth}{0pt}}
\includegraphics[width=1.0\linewidth]{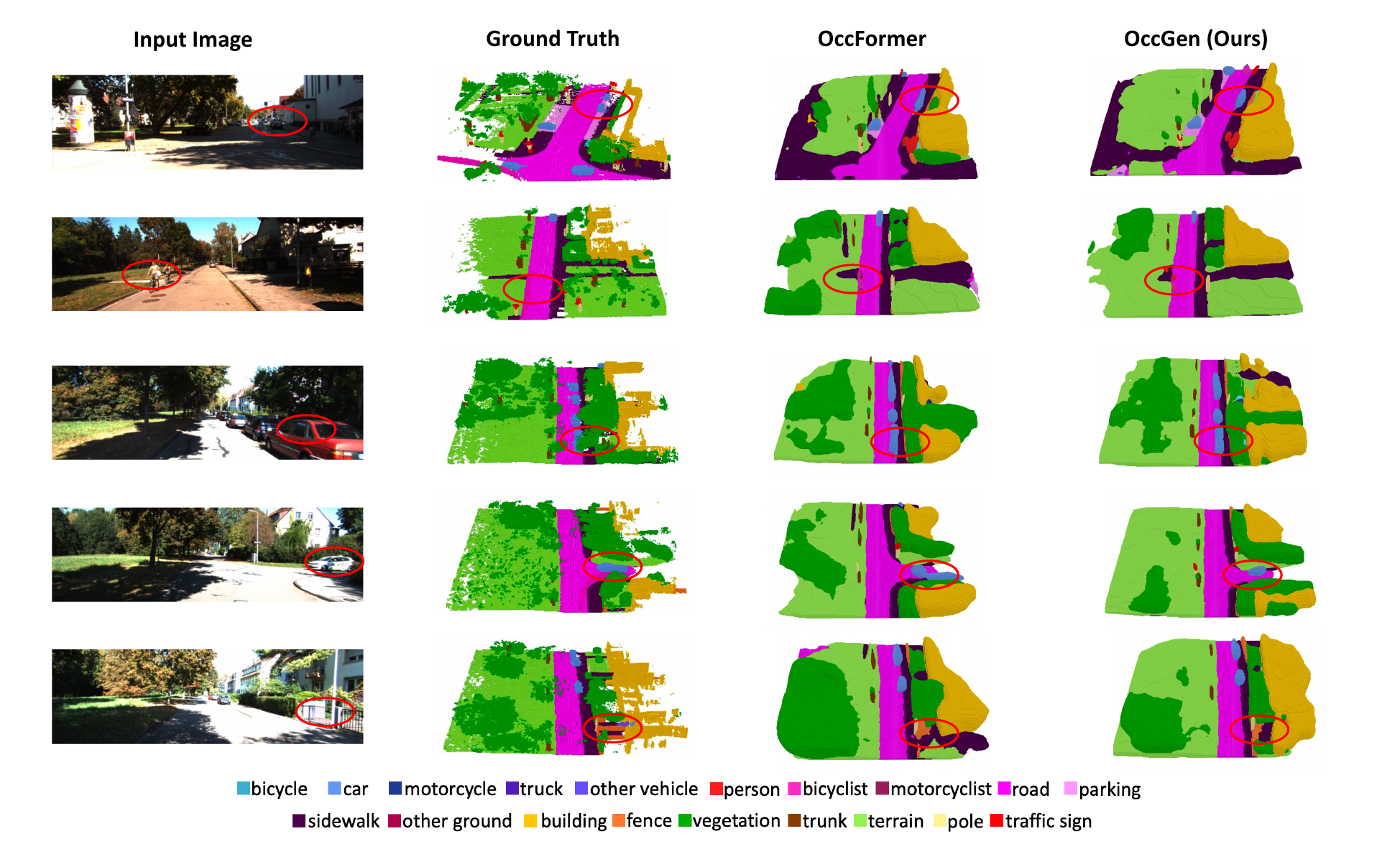}
% \vspace{-1.0em}
\end{center}
   \caption{Qualitative results of semantic Scene Completion on SemanticKITTI~\cite{behley2019semantickitti} validation set.
   The leftmost column shows the input image, the following three columns visualize the results from the ground truth, OccFormer\cite{Zhang_2023_ICCV}, and Our OccGen.}
\label{fig_vis_kitti}
\vspace{-1em}
\end{figure}

\begin{figure}
\begin{center}
%\fbox{\rule{0pt}{2in} \rule{0.9\linewidth}{0pt}}
\includegraphics[width=1.0\linewidth]{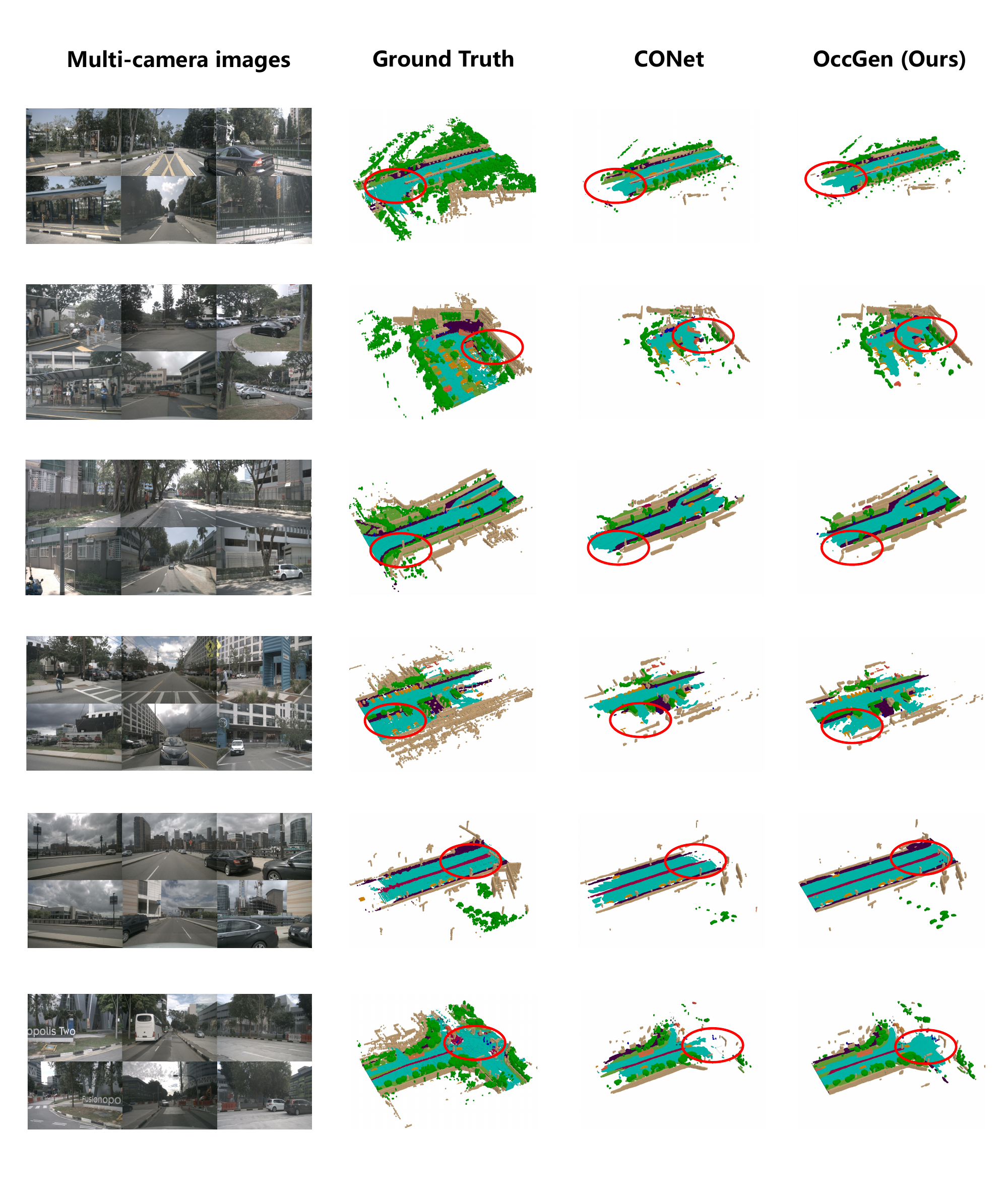}
% \vspace{-1.5em}
\end{center}
   \caption{Qualitative results of the 3D semantic occupancy predictions on nuScenes-Occupancy. The leftmost column shows the input surrounding images, the following three columns visualize the 3D semantic occupancy results from the ground truth, CONet\cite{wang2023openoccupancy}, and Our OccGen. The regions highlighted by red circles indicate that these areas have obvious differences (better viewed when zoomed in).}
\label{fig_ex_compare}
% \vspace{-1em}
\end{figure}

\begin{figure*}
\begin{center}
%\fbox{\rule{0pt}{2in} \rule{0.9\linewidth}{0pt}}
\includegraphics[width=1.0\linewidth]{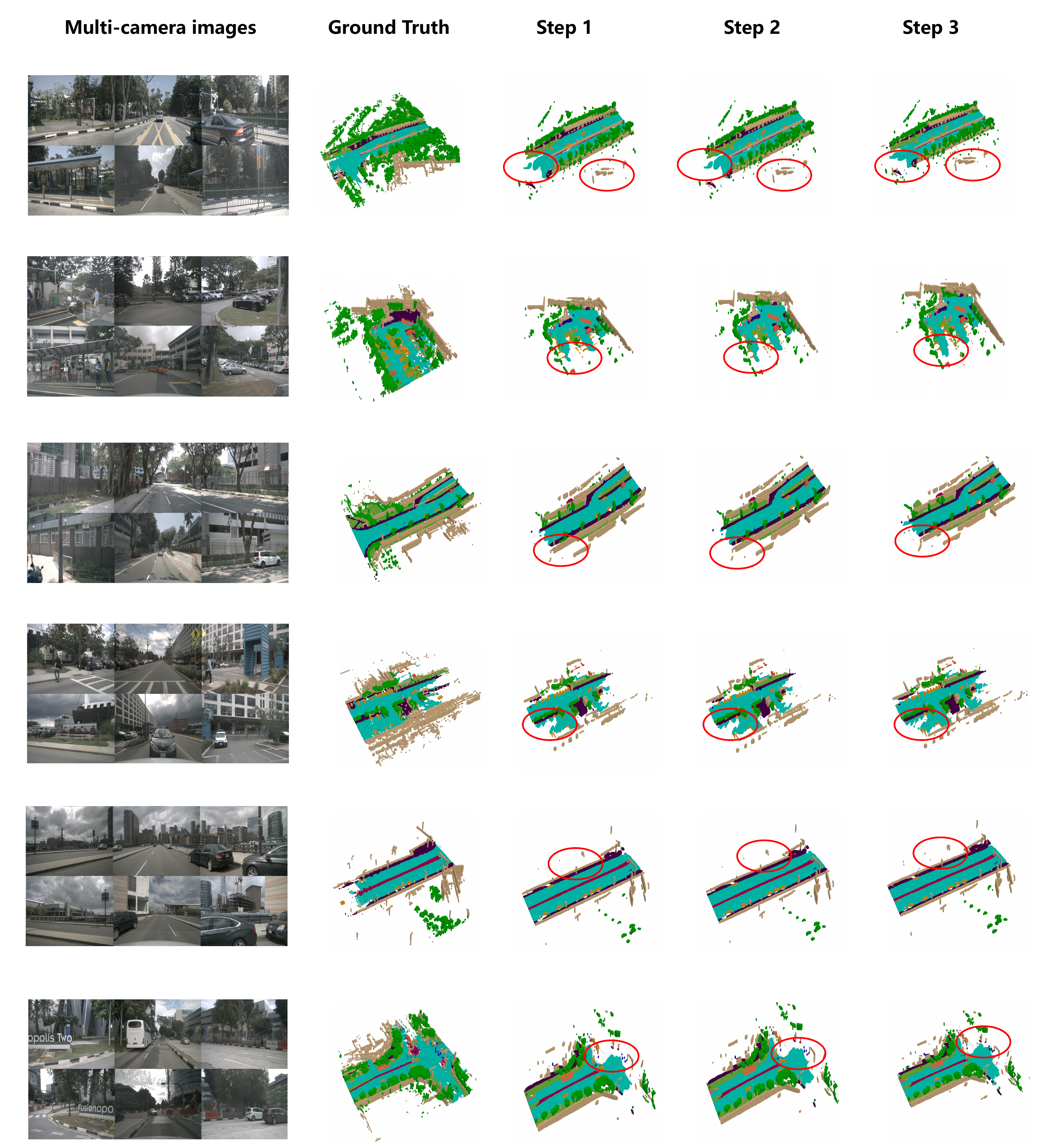}
% \vspace{-1.5em}
\end{center}
   \caption{The predicted occupancy results on the different steps of OccGen on nuScenes-Occupancy. The leftmost column shows the input surrounding images, the following four columns visualize the 3D semantic occupancy results from the ground truth, step 1, step 2, and step 3. The regions highlighted by red circles indicate that these areas have obvious differences (better viewed when zoomed in).}
\label{fig_ex_steps}
% \vspace{-1em}
\end{figure*}

\begin{figure*}
\begin{center}
%\fbox{\rule{0pt}{2in} \rule{0.9\linewidth}{0pt}}
\includegraphics[width=0.95\linewidth]{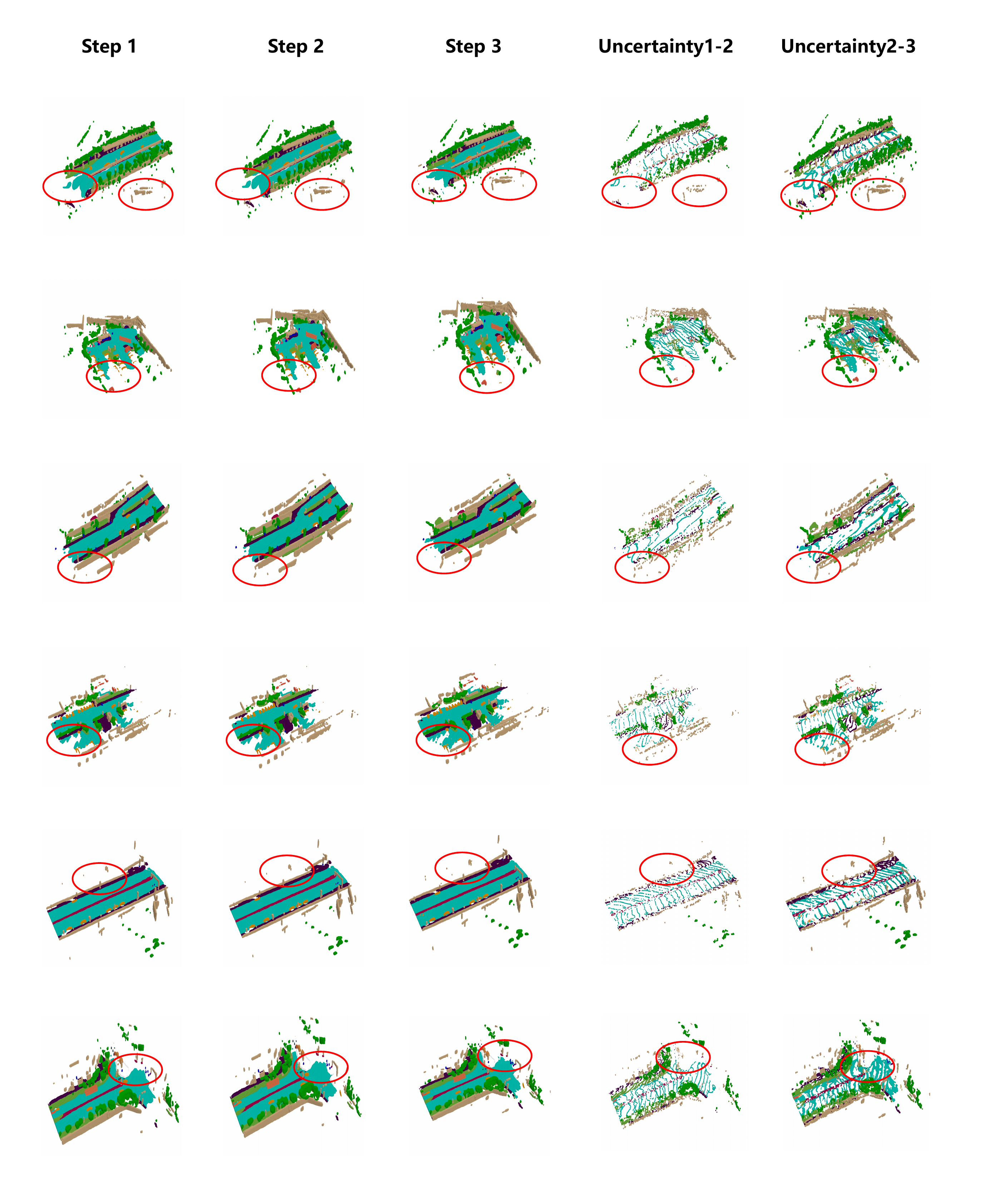}
% \vspace{-1.5em}
\end{center}
   \caption{The uncertainty estimates between different steps of OccGen on nuScenes-Occupancy. The left three columns show the predicted 3D semantic occupancy results from step 1, step 2, and step 3. The ``Uncertainty 1-2" and ``Uncertainty 2-3" represent the high estimated uncertainty voxels from step one to step two and from step two to step three, respectively. The regions highlighted by red circles indicate that these areas have obvious differences (better viewed when zoomed in).}
\label{fig_ex_uncertainty}
% \vspace{-1em}
\end{figure*}

\paragraph{More visualization.} 
We visualize the predicted results of OccFormer~\cite{Zhang_2023_ICCV} and Our OccGen on SemanticKITTI~\cite{behley2019semantickitti} in Fig~\ref{fig_vis_kitti}. We can observe that OccGen produces more reasonable results than OccFormer~\cite{Zhang_2023_ICCV}.
In Fig.~\ref{fig_ex_compare}, we visualize the predicted results of 3D semantic occupancy on nuScenes-Occupancy from CONet \cite{wang2023openoccupancy} and our proposed OccGen. It is evident that the ``drivable surface'' and ``sidewalk'' regions predicted by our OccGen exhibit superior continuity and integrity. This results in a significant reduction in the number of void areas compared to the previous SOTA CONet \cite{wang2023openoccupancy}. In Fig.~\ref{fig_ex_steps}, we visualize the predicted results of 3D occupancy of different sampling steps. We observe that the results of the third step have more complete geometric structure and semantic information compared with the generated results of the first step. In Fig.~\ref{fig_ex_uncertainty}, we note that the uncertainty maps of different steps clearly show that the proposed OccGen can iteratively refine the occupancy in a coarse-to-fine manner.

\section{Broader Impact Statement and Limitations}
% \vspace{-0.2cm}
This paper studies a generative model for 3D occupancy semantic prediction and does not see potential privacy-related issues. Nevertheless, the deployment of a model that is biased toward the training data may introduce significant safety concerns and potential risks when utilized in real-world applications. This research is simple yet effective, which may inspire the community to produce follow-up generative studies for 3D occupancy.

% \newpage

% \newpage
% \newpage

% \newpage

\end{document}